\pdfoutput=1
\documentclass[10pt]{article}
\usepackage{samaya}
\titleformat{\paragraph}[runin]{\normalfont\bfseries\color{SamayaInk}}{}{0pt}{}
\usepackage[numbers,sort&compress]{natbib}
\usepackage{xurl}
\usepackage{pgfplots}
\pgfplotsset{compat=1.16}
\usetikzlibrary{arrows.meta,positioning,fit,backgrounds,calc}
\usepackage{algorithm}
\usepackage{algpseudocode}
\algrenewcommand\algorithmicrequire{\textbf{Input:}}
\algrenewcommand\algorithmicensure{\textbf{Output:}}
\definecolor{figAcq}{HTML}{2A78D6}\definecolor{figFil}{HTML}{EB6834}\definecolor{figCon}{HTML}{1BAF7A}
\definecolor{figInk}{HTML}{0B0B0B}\definecolor{figInk2}{HTML}{52514E}\definecolor{figMute}{HTML}{8A8984}
\definecolor{figSurf}{HTML}{F0EFEC}\definecolor{figSurf2}{HTML}{FAFAF8}
\definecolor{figAcqBg}{HTML}{E4EEFB}\definecolor{figFilBg}{HTML}{FCE9E0}\definecolor{figConBg}{HTML}{E0F5EE}
\tikzset{
  figfont/.style={font=\sffamily\fontsize{7.4}{9}\selectfont, text=figInk},
  figfontsm/.style={font=\sffamily\fontsize{6.6}{8}\selectfont, text=figInk2},
  fbox/.style={figfont, draw=figMute, line width=0.5pt, rounded corners=1.6pt, fill=figSurf2,
               inner xsep=5pt, inner ysep=3.6pt, align=center, minimum height=15pt},
  facq/.style={fbox, draw=figAcq, fill=figAcqBg}, ffil/.style={fbox, draw=figFil, fill=figFilBg},
  fcon/.style={fbox, draw=figCon, fill=figConBg}, fplain/.style={fbox, draw=figMute, fill=figSurf},
  farr/.style={-{Stealth[length=4pt,width=3pt]}, draw=figInk2, line width=0.6pt},
  farrd/.style={-{Stealth[length=4pt,width=3pt]}, draw=figMute, line width=0.55pt, dash pattern=on 2pt off 1.6pt},
  flab/.style={figfontsm, fill=white, inner sep=1.4pt},
  fgroup/.style={draw=figMute, line width=0.5pt, rounded corners=2.5pt, dash pattern=on 2pt off 1.6pt, inner sep=6pt},
}

\usepackage{placeins}

\newcommand{\sigp}{\sigma_p}
\newcommand{\ci}[2]{[#1,\,#2]}
\usepackage{etoolbox}

\AtBeginEnvironment{tabular}{\let\ci\cismall}
\newcommand{\shape}[2]{#1{\times}#2}

\definecolor{cUnif}{HTML}{2A78D6}
\definecolor{cOurs}{HTML}{EB6834}
\definecolor{cDapo}{HTML}{3A3A3A}\definecolor{cZero}{HTML}{D03B3B}

\title{Thompson-Sampled Curriculum for GRPO:\\Improving the Accuracy-Rollout Efficiency Frontier}
\author{O. Ozan Koyluoglu\\[4pt]
\samayaaffil{Samaya AI}}
\date{}
\samayapreprint{Samaya AI Technical Report}
\begin{document}
\begin{samayatitle}
Group Relative Policy Optimization (GRPO) learns from a prompt only when the answers sampled
for it receive different rewards. On standard math datasets, about half of the generated
answers (rollouts) fall in groups whose rewards are all equal, called zero-variance groups, and
the compute spent on them produces no gradient. Existing remedies trade accuracy against cost.
Filtering the pool once before training cannot follow the model as it learns, and discarding
zero-variance groups after generation, as in DAPO, pays for the rollouts it discards. To avoid
this waste, we develop and evaluate the Thompson-sampled curriculum (TSC). Before generation, TSC predicts the prompts that will
produce varied rewards, and it updates this prediction during training. We map the frontier of
accuracy against training efficiency on a math task for TSC and uniform sampling across batch configurations
and rollout budgets, with the learning rate tuned for each method. TSC increases the accuracy
gain from training by about 32\% while removing 55--96\% of the zero-variance groups, without
generating extra rollouts. Uniform sampling stops improving as the budget grows, while TSC keeps
improving, so TSC is the more efficient method when the target accuracy is high. DAPO needs
about twice as many rollouts as TSC to reach the same accuracy. Selecting prompts once before
training recovers half to two thirds of the gain of TSC. We also study whether TSC generalizes beyond
math. On a code generation task where a program is rewarded only if it passes all unit
tests, TSC increases the accuracy gain from training by about 28\%. It also reaches the best
held-out accuracy of uniform sampling with 63\% of the rollouts. On a research agent task with
graded rewards, TSC reaches the final score of uniform sampling with fewer than half of the
rollouts.
\end{samayatitle}
\suppressfloats[t]

\section{Introduction}\label{sec:intro}

Reinforcement learning with verifiable rewards trains a language model on prompts whose
answers a program can check. The model answers a prompt, and a verifier scores the answer,
either as correct or incorrect, or with a graded score between the two. The model is then
updated to make answers with a higher score more likely. Group Relative
Policy Optimization (GRPO) \citep{shao2024deepseekmath} is a widely used algorithm for this
setting and was used to train DeepSeek-R1 \citep{deepseek2025r1}. For each prompt, GRPO
generates a group of $G$ answers and compares the reward of each answer with the mean reward of its group.
Answers that score above the group mean are reinforced, and answers below it are discouraged.

This comparison within a group fails when all $G$ answers receive the same reward. Every
reward then equals the group mean, and the prompt contributes nothing to the update. We call
such a group a \emph{zero-variance group}. Consider a problem that the model cannot
yet solve. GRPO generates eight answers, the verifier marks all eight wrong, and the tokens
spent on them change no parameter. The same happens for a problem that the model always
solves. On the DeepScaleR mathematics pool \citep{luo2025deepscaler},
the untrained Qwen3-1.7B model answers 51.4\% of the 37{,}970 prompts correctly either zero or eight times
out of eight. Uniform sampling therefore spends about half of its generation on prompts that
return no gradient, a cost that several recent papers have noted
\citep{yu2025dapo,xu2025spo,xiong2025reinforceada}.

There are two common remedies. The first filters the pool by difficulty, once or at a few
points in training \citep{an2025polaris}. Between these points, the filter cannot follow the
model as its abilities change. The second generates answers
first and then discards the zero-variance groups, generating more until the batch is full
\citep{yu2025dapo}. This second remedy follows the model, but the discarded rollouts increase the cost of
generation.

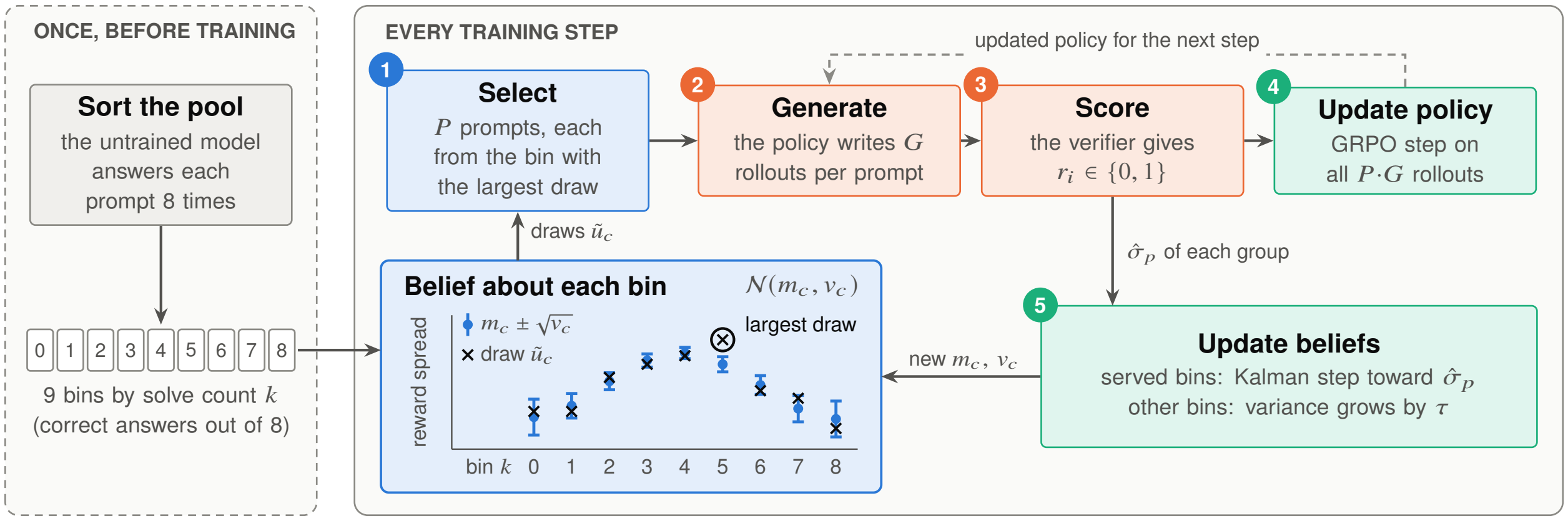
\begin{figure}[t]
\centering
\def\tsT#1{{\sffamily\bfseries\fontsize{8.4}{10}\selectfont #1}}%
\def\tsS#1{{\sffamily\fontsize{7}{8.6}\selectfont\color{figInk2}#1}}%
\begin{tikzpicture}[figfont, x=1cm, y=1cm,
  stp/.style={fbox, line width=0.7pt, rounded corners=2.4pt, inner xsep=5pt, inner ysep=5pt,
              minimum height=1.9cm, text width=2.42cm, align=center},
  bdg/.style={circle, fill=#1, text=white, inner sep=0pt, minimum size=10.5pt,
              font=\sffamily\bfseries\fontsize{6.8}{7}\selectfont},
  zone/.style={rounded corners=4pt, line width=0.5pt},
  zlab/.style={font=\sffamily\bfseries\fontsize{6.6}{8}\selectfont, text=figInk2},
  flow/.style={-{Stealth[length=5pt,width=4pt]}, draw=figInk2, line width=0.8pt},
  flowd/.style={-{Stealth[length=5pt,width=4pt]}, draw=figMute, line width=0.7pt, dash pattern=on 2.4pt off 1.8pt},
  note/.style={font=\sffamily\fontsize{6.6}{8}\selectfont, text=figInk2, inner sep=1.5pt}]
\fill[zone, figSurf2] (-0.18,-3.62) rectangle (3.12,1.78);
\draw[zone, figMute, dash pattern=on 2.4pt off 1.8pt] (-0.18,-3.62) rectangle (3.12,1.78);
\node[zlab, anchor=west] at (0.0,1.50) {ONCE, BEFORE TRAINING};
\fill[zone, figSurf2] (3.52,-3.62) rectangle (16.32,1.78);
\draw[zone, figMute] (3.52,-3.62) rectangle (16.32,1.78);
\node[zlab, anchor=west] at (3.72,1.50) {EVERY TRAINING STEP};
\node[stp, fplain, text width=2.42cm] (base) at (1.47,0.20)
  {\tsT{Sort the pool}\\[1pt]\tsS{the untrained model answers each prompt 8~times}};
\draw[flow] (base.south) -- (1.47,-1.60);
\foreach \k in {0,...,8} {
  \node[draw=figMute, fill=white, line width=0.5pt, rounded corners=1pt, inner sep=0pt,
        minimum width=0.275cm, minimum height=0.44cm,
        font=\sffamily\fontsize{6.2}{7}\selectfont, text=figInk2] at (0.19+\k*0.32,-1.87) {\k};
}
\node[align=center, text width=2.9cm] at (1.47,-2.52) {\tsS{9 bins by solve count $k$\\ (correct answers out of 8)}};
\node[stp, facq] (sel) at (5.25,0.35)
  {\tsT{Select}\\[1pt]\tsS{$P$ prompts, each from the bin with the largest draw}};
\node[stp, ffil] (gen) at (8.55,0.35)
  {\tsT{Generate}\\[1pt]\tsS{the policy writes $G$ rollouts per prompt}};
\node[stp, ffil] (ver) at (11.55,0.35)
  {\tsT{Score}\\[1pt]\tsS{the verifier gives \mbox{$r_i\in\{0,1\}$}}};
\node[stp, fcon] (grpo) at (14.65,0.35)
  {\tsT{Update policy}\\[1pt]\tsS{GRPO step on all $P{\cdot}G$ rollouts}};
\node[bdg=figAcq] at (sel.north west) {1};
\node[bdg=figFil] at (gen.north west) {2};
\node[bdg=figFil] at (ver.north west) {3};
\node[bdg=figCon] at (grpo.north west) {4};
\draw[flow] (sel) -- (gen);
\draw[flow] (gen) -- (ver);
\draw[flow] (ver) -- (grpo);
\fill[figAcqBg, rounded corners=2.4pt] (3.80,-3.38) rectangle (9.10,-0.92);
\draw[figAcq, line width=0.7pt, rounded corners=2.4pt] (3.80,-3.38) rectangle (9.10,-0.92);
\node[anchor=west] at (3.92,-1.19) {\tsT{Belief about each bin}};
\node[anchor=east] at (9.0,-1.19) {\tsS{$\mathcal N(m_c, v_c)$}};
\draw[figInk2, line width=0.45pt] (4.55,-2.92) -- (8.85,-2.92);
\draw[figInk2, line width=0.45pt] (4.55,-2.92) -- (4.55,-1.50);
\node[note, rotate=90] at (4.22,-2.21) {reward spread};
\node[font=\sffamily\fontsize{6.6}{8}\selectfont, text=figInk2, inner sep=0pt, anchor=base east] at (5.20,-3.18) {bin $k$};
\foreach \k/\m/\s/\d in {0/0.24/0.19/0.30, 1/0.36/0.13/0.30, 2/0.62/0.09/0.66, 3/0.84/0.07/0.80,
                         4/0.92/0.06/0.89, 5/0.80/0.08/1.06, 6/0.58/0.10/0.52, 7/0.33/0.14/0.44, 8/0.22/0.19/0.12} {
  \pgfmathsetmacro{\xx}{5.42+\k*0.40}
  \draw[figAcq, line width=0.9pt] (\xx,-2.82+\m-\s) -- (\xx,-2.82+\m+\s);
  \draw[figAcq, line width=0.9pt] (\xx-0.06,-2.82+\m-\s) -- (\xx+0.06,-2.82+\m-\s);
  \draw[figAcq, line width=0.9pt] (\xx-0.06,-2.82+\m+\s) -- (\xx+0.06,-2.82+\m+\s);
  \fill[figAcq] (\xx,-2.82+\m) circle (1.5pt);
  \draw[figInk, line width=0.7pt] (\xx-0.055,-2.82+\d-0.055) -- (\xx+0.055,-2.82+\d+0.055);
  \draw[figInk, line width=0.7pt] (\xx-0.055,-2.82+\d+0.055) -- (\xx+0.055,-2.82+\d-0.055);
  \node[font=\sffamily\fontsize{6.6}{8}\selectfont, text=figInk2, inner sep=0pt, anchor=base] at (\xx,-3.18) {\k};
}
\draw[figInk, line width=0.8pt] (7.42,-1.76) circle (3.4pt);
\node[note, anchor=west, text=figInk] at (7.60,-1.62) {largest draw};
\fill[figAcq] (4.72,-1.60) circle (1.5pt);
\draw[figAcq, line width=0.9pt] (4.72,-1.72) -- (4.72,-1.48);
\node[note, anchor=west] at (4.80,-1.60) {\mbox{$m_c\pm\sqrt{v_c}$}};
\draw[figInk, line width=0.7pt] (4.665,-1.975) -- (4.775,-1.865);
\draw[figInk, line width=0.7pt] (4.665,-1.865) -- (4.775,-1.975);
\node[note, anchor=west] at (4.80,-1.92) {draw $\tilde u_c$};
\node[stp, fcon, text width=4.9cm, minimum height=1.5cm] (kal) at (13.42,-2.15)
  {\tsT{Update beliefs}\\[1pt]\tsS{served bins: Kalman step toward $\hat\sigma_p$\\
   other bins: variance grows by $\tau$}};
\node[bdg=figCon] at (kal.north west) {5};
\draw[flow] (2.92,-1.87) -- (3.80,-1.87);
\draw[flow] (5.25,-0.92) -- (sel.south);
\node[note, anchor=west] at (5.33,-0.62) {draws $\tilde u_c$};
\draw[flow] (ver.south) -- (11.55,-1.40);
\node[note, anchor=west] at (11.63,-0.85) {$\hat\sigma_p$ of each group};
\draw[flow] (kal.west) -- (9.10,-2.15);
\node[note, anchor=south] at (9.98,-2.13) {new $m_c,\,v_c$};
\draw[flowd] (grpo.north) -- ++(0,0.34) -| (gen.north);
\node[note, fill=figSurf2, inner sep=2pt] at (11.6,1.40) {updated policy for the next step};
\end{tikzpicture}
\caption{\textbf{The Thompson-sampled curriculum (TSC) in one GRPO training step.} Before
training, the untrained model answers every prompt eight times, and the pool is sorted into
nine bins by the number of correct answers. TSC keeps a Gaussian belief about the reward
spread of each bin. At each step, it (1)~draws once from every belief and takes a prompt from
the bin with the largest draw, $P$ times, (2)~lets the policy generate $G$ rollouts per prompt,
(3)~scores them, and (4)~applies the usual GRPO update to all rollouts. It then (5)~moves the
belief of each served bin toward the reward spread $\hat\sigma_p$ that its groups produced,
and makes the beliefs of the other bins less certain so that they are tried again later. Blue
marks selection, orange marks generation and scoring, and green marks the updates.}
\label{fig:tsc}
\end{figure}

A third option is to predict, before generating, which prompts will produce varied rewards,
and to update that prediction during training. We study a simple method of this kind, which
we call the \emph{Thompson-sampled curriculum} (TSC). TSC sorts the prompts of the pool into bins
by the number of correct answers that the untrained model gives to each prompt (with graded rewards,
the bins are quantiles of each prompt's mean reward). Our main experiments use nine bins, for
zero to eight correct answers out of eight. For each bin, TSC keeps an estimate of the
reward spread (the standard deviation of the rewards within a group) that the bin's prompts currently produce. At each step, TSC uses Thompson sampling
\citep{thompson1933,russo2014learning} to choose the bins with the largest estimated spread
(Figure~\ref{fig:tsc}, Algorithm~\ref{alg:tsc}). Every generated rollout is used for training.
Under binary rewards, the reward spread of a group is proportional to the total magnitude of
its GRPO advantages (Proposition~\ref{prop:identity}), so TSC directs generation to the
prompts that contribute most to the update. We study when such a curriculum improves
on uniform sampling and on other selection methods, and by how much.

Answering this question needs a careful comparison, for two reasons. First, a GRPO step has a
budget of rollouts that can be divided into many prompts with few answers each, or few prompts
with many answers each. Prior work studies how to spend a rollout budget across prompts
\citep{kim2026pilotcommit,wang2026wheretospend}, but, as far as we can tell, each selection
method keeps the number of prompts per step fixed, so none trades prompts per step against
answers per prompt at a fixed budget. Second, TSC changes the size of the
update, so the best learning rate for TSC need not be the best one for uniform sampling. We
therefore train Qwen3-1.7B with both methods in ten batch configurations, tune the learning
rate for each method in each configuration, and run two seeds per setting. This gives 84
combinations of configuration, method and learning rate, and 168 runs.

\begin{figure}[t]
\centering
\begin{minipage}[t]{.49\linewidth}\centering
\begin{tikzpicture}
\begin{axis}[
  width=\linewidth, height=6.0cm,
  title={Uniform plateaus, TSC keeps improving},
  title style={font=\footnotesize\bfseries, yshift=-3pt},
  xmode=log, log basis x=2,
  xmin=3.2, xmax=47, ymin=37, ymax=43.7,
  xtick={4,8,16,32}, xticklabels={4k,8k,16k,32k},
  ytick={37,38,39,40,41,42,43},
  xlabel={rollout budget (rollouts in 250 steps)},
  ylabel={best avg@16 at this budget or less},
  legend pos=north west, legend cell align=left, legend style={font=\scriptsize},
  tick label style={font=\small}, label style={font=\small},
  grid=major, grid style={line width=.1pt, draw=gray!18},
]
\addplot[fill=figFilBg, draw=none, forget plot] coordinates {(16,41.64) (16,42.25) (32,42.94) (32,41.64)} -- cycle;
\addplot[gray, dashed, thin, forget plot] coordinates {(3.2,37.54) (47,37.54)};
\node[font=\scriptsize, text=gray, anchor=south east] at (axis cs:46,37.54) {untrained model, 37.5};
\addplot[cUnif, very thick, mark=*, mark size=2.3pt] coordinates {(4,40.95) (8,40.95) (16,41.64) (32,41.64)};
\addplot[cOurs, very thick, mark=square*, mark size=2.3pt] coordinates {(4,39.59) (8,40.77) (16,42.25) (32,42.94)};
\draw[<->, >=stealth, figInk2, thin] (axis cs:32,41.84) -- (axis cs:32,42.74);
\node[font=\scriptsize, text=figInk2, anchor=east] at (axis cs:31,42.29) {$+3.1\%$};
\node[font=\scriptsize\bfseries, anchor=west] at (axis cs:34,42.94) {42.9};
\node[font=\scriptsize, anchor=west] at (axis cs:34,41.64) {41.6};
\node[font=\scriptsize, text=figInk2, anchor=north, align=center] at (axis cs:22.6,41.5) {uniform stops\\ improving at 41.6};
\legend{uniform sampling, TSC}
\end{axis}
\end{tikzpicture}
\end{minipage}\hfill
\begin{minipage}[t]{.49\linewidth}\centering
\begin{tikzpicture}
\begin{axis}[
  width=\linewidth, height=6.0cm,
  title={At high accuracy, TSC is more efficient},
  title style={font=\footnotesize\bfseries, yshift=-3pt},
  xmode=log, log basis x=2,
  xmin=1.05, xmax=12.5, ymin=38.9, ymax=43.7,
  xtick={1.25,2.5,5,10}, xticklabels={1.25,2.5,5,10},
  ytick={39,40,41,42,43},
  xlabel={avg@16 per 1{,}000 rollouts generated},
  ylabel={avg@16 (held-out DeepScaleR)},
  tick label style={font=\small}, label style={font=\small},
  grid=major, grid style={line width=.1pt, draw=gray!18},
]
\addplot[fill=figConBg, draw=none, forget plot] coordinates {(1.05,41.64) (12.5,41.64) (12.5,43.7) (1.05,43.7)} -- cycle;
\node[font=\scriptsize\sffamily, text=figCon!70!black, anchor=north east] at (axis cs:12.1,43.62) {above every uniform setting};
\addplot[only marks, mark=o, mark size=1.7pt, cUnif, forget plot] coordinates {(4.98,39.84) (2.53,40.48) (2.515,40.24) (1.2772,40.87) (1.2947,41.43) (1.2953,41.45)};
\addplot[only marks, mark=square, mark size=1.6pt, cOurs, forget plot] coordinates {(5.0613,40.49) (2.57,41.12) (2.5819,41.31) (1.3219,42.30) (1.3203,42.25) (1.3291,42.53)};
\addplot[cUnif, very thick, mark=*, mark size=2.3pt] coordinates {(1.2997,41.59) (2.6025,41.64) (5.0875,40.70) (10.2375,40.95)};
\addplot[cOurs, very thick, mark=square*, mark size=2.3pt] coordinates {(1.3419,42.94) (2.6406,42.25) (5.0963,40.77) (9.8975,39.59)};
\node[font=\scriptsize, text=gray, anchor=south] at (axis cs:1.32,38.93) {32k};
\node[font=\scriptsize, text=gray, anchor=south] at (axis cs:2.62,38.93) {16k};
\node[font=\scriptsize, text=gray, anchor=south] at (axis cs:5.09,38.93) {8k};
\node[font=\scriptsize, text=gray, anchor=south] at (axis cs:10.07,38.93) {4k};
\draw[figInk2, thin] (axis cs:2.82,42.33) -- (axis cs:3.5,42.72);
\node[font=\scriptsize, text=figInk2, anchor=west, align=left] at (axis cs:3.5,42.78) {TSC with 16k rollouts: 42.25};
\end{axis}
\end{tikzpicture}
\end{minipage}
\caption{\textbf{Accuracy and rollout efficiency of TSC and uniform sampling} (Qwen3-1.7B,
250 steps, ten batch configurations, two-seed means, each method at its best learning rate).
Left: the best accuracy (avg@16, the fraction of 16 rollouts per held-out problem that are correct) that each method reaches with a given number of generated rollouts or
fewer. Uniform sampling stops improving at 41.6\%. TSC reaches 42.3\% with 16{,}000 rollouts
and 42.9\% with 32{,}000, a relative gain of 3.1\% over uniform sampling. Right: every configuration by its accuracy and by its accuracy per
1{,}000 generated rollouts. The four columns of points are the four budgets, which are
labeled at the bottom. Filled markers joined by a line are the best configuration of each
method at each budget, and hollow markers are the other configurations. No uniform setting
lies in the shaded band. TSC enters the band with 16{,}000 rollouts, half of the largest
budget.}
\label{fig:headline}
\end{figure}
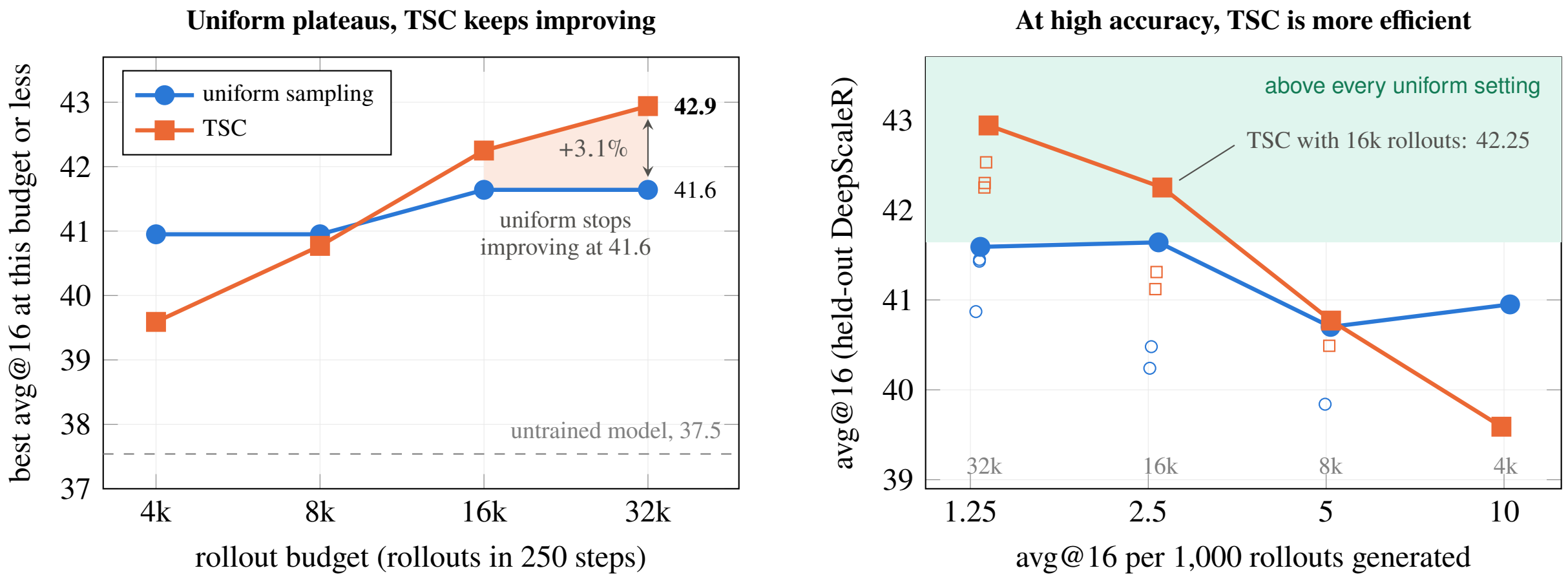

Figure~\ref{fig:headline} summarizes the result on the math task. TSC removes most zero-variance groups in
every configuration. TSC raises the best held-out accuracy from 41.6\% to 42.9\%, a relative
gain of 3.1\%. The untrained model scores 37.5\%, so training adds 4.1 points with uniform
sampling and 5.4 points with TSC, and TSC increases the gain from training by 31.7\%. The benefit depends on the budget. Uniform sampling stops
improving at 16{,}000 rollouts, and TSC keeps improving up to the largest budget we tested.
The right panel of Figure~\ref{fig:headline} measures efficiency as accuracy per 1{,}000
generated rollouts. In the high-accuracy regime, TSC reaches a higher accuracy than uniform
sampling at the same efficiency. With 16{,}000 rollouts, TSC reaches 42.3\%, which no uniform
setting reaches with any budget we tested. TSC is therefore the more efficient choice when the
target accuracy is high (Figure~\ref{fig:master} in Appendix~\ref{app:grid} shows every run).
In the following sections, we detail what the gain consists of (Section~\ref{sec:coverage})
and what predicts it (Section~\ref{sec:diversity}), and we compare TSC with other selection
methods (Section~\ref{sec:comparators}) and test it on other model sizes
(Section~\ref{sec:second-model}) and on other tasks (Sections~\ref{sec:code}
and~\ref{sec:agent}).

\paragraph{Contributions.}
\begin{enumerate}\itemsep2pt
\item \textbf{A controlled comparison of a pre-generation curriculum with uniform sampling.}
Across ten batch configurations, each method at its own best learning rate, TSC improves
accuracy in seven configurations and raises the best accuracy by 3.1\% in relative terms and
the gain from training by 31.7\%, without generating any extra rollout. The gain grows with the
number of rollouts per step. The two methods prefer different learning rates and opposite
batch configurations, so we tune both for each method and compare the methods at their own
best settings (Sections~\ref{sec:uniform}--\ref{sec:frontier}).
\item \textbf{Reliability gain.} TSC increases the share of problems that the model solves
in all 16 attempts by 0.89 points, mostly on problems of medium difficulty
(Section~\ref{sec:coverage}).
\item \textbf{A predictor of when TSC helps.} The variety of difficulty within a
step predicts how much TSC can gain. We measure this variety as the entropy of the
difficulty bins that TSC serves in one step. Across the ten batch configurations, the gain
rises by about 5.1 points per bit of entropy ($r=0.88$; Section~\ref{sec:diversity}).
\item \textbf{Comparison with other selection methods and other settings.} At the same
configuration, selecting prompts once before training recovers half to two thirds of TSC's gain. DAPO
reaches the same accuracy while generating 2.1 times as many rollouts. The accuracy gains and
the rollout savings also hold on the smaller and the larger model we tested, Qwen3-0.6B and
Qwen3-4B. On two other tasks, TSC also saves rollouts. On a code task, it increases the
accuracy gain from training by about 28\% and reaches the best held-out accuracy of uniform
sampling with 63\% of the rollouts. On an agent task with graded rewards, it reaches the
final score of uniform sampling with 45\% of the rollouts
(Sections~\ref{sec:comparators}--\ref{sec:agent}).
\end{enumerate}

\section{Related Work}\label{sec:related}

Methods that deal with zero-variance groups differ in when they decide which prompts to use.
This timing largely determines their cost, and we organize prior work by it.

\paragraph{Deciding before training.} Ordering training data from easy to hard has a long
history \citep{bengio2009curriculum}, and E2H \citep{parashar2025e2h} applies such schedules to
RL for language-model reasoning. A related practice is filtering by difficulty. POLARIS
\citep{an2025polaris} removes the prompts that the model being trained solves in all eight of
eight attempts before training, and at the end of each training stage removes the prompts
whose accuracy exceeds 0.9. Between stages, the pool does not change as the model improves. We compare TSC with
offline filtering and with a static weighting of the bins in Section~\ref{sec:comparators}.

\paragraph{Deciding after generating.} DAPO \citep{yu2025dapo} generates answers for more
prompts than it needs, drops the groups whose rewards all agree, and refills the batch. VCRL
\citep{jiang2025vcrl} drops prompts whose reward variance is below a threshold and refills the
batch from a replay bank of earlier high-variance prompts. Reinforce-Ada
\citep{xiong2025reinforceada} keeps sampling a prompt in rounds until it has enough correct
and incorrect answers or reaches a cap, and then reduces the collected answers to a balanced
group.
StarPO-S \citep{wang2025ragen} keeps the prompts with the highest reward standard deviation.
Online difficulty filtering \citep{bae2025onlinefiltering} keeps the prompts whose observed
solve rate lies in a band around one half.
Pilot-Commit \citep{kim2026pilotcommit} spends some rollouts to estimate each prompt's success
rate and spends the rest on prompts in a middle band. It reports reaching a target accuracy
with 2.3--4.0$\times$ fewer rollouts than DAPO. LILO \citep{foster2025learnability} scores
prompts by $p(1-p)$, where $p$ is the prompt's success rate, estimated from a few rollouts on a larger candidate pool, and trains on
the prompts with the highest score. All of these methods measure the reward variance of a prompt only after
generating rollouts for it, which is the cost that TSC tries to avoid.

\paragraph{Changing the objective.} Other work changes how the update uses a group instead of
which prompts are sampled. SPO \citep{xu2025spo} replaces the group baseline.
\citet{le2025zerovariance} give zero-variance prompts a learning signal through
entropy-guided advantage shaping. \citet{nie2026starvation} analyze why subtracting the group
mean removes the gradient under binary rewards.

\paragraph{Deciding before generating.} The work closest to ours chooses prompts before
generation. GRESO \citep{zheng2025greso} skips a prompt with a probability that grows with the
number of consecutive recent groups in which it produced zero variance. SPaCe, first released as SPaRFT \citep{do2025sparft},
clusters prompts by sentence embedding and difficulty. At each step, it uses Thompson sampling
to choose one cluster, draws the whole batch from it, and prefers clusters where the model's
reward is low, with a penalty for clusters that stop improving. AdaRFT \citep{shi2025adarft}
adjusts a target difficulty to keep the model's accuracy near a chosen level. KGPS
\citep{zhu2026kgps} keeps a Kalman filter for each prompt and selects the prompts with the
highest expected $p(1-p)$, where $p$ is the solve probability. MoPPS \citep{qu2025mopps}
draws each prompt's success rate from a Beta posterior and keeps the prompts whose draw lies
closest to a target near one half. Actor-Curator
\citep{gu2026actorcurator} treats problem selection as a non-stationary bandit and trains a
neural curator to select the problems that most improve the policy. Contextual Rollout Bandits
\citep{lu2026contextualbandits} use a contextual bandit to decide which rollouts to use within
a group and which earlier rollouts to reuse.

A cluster-level Thompson sampler on a variance-like signal is therefore already part of this
literature, and we do not claim the sampler as a contribution. TSC differs from the per-prompt
methods (GRESO, KGPS, MoPPS) in one practical respect. A per-prompt estimate improves only
when its prompt is served again, and at our budgets, the average prompt is served between 0.03
and 0.21 times in a 250-step run. TSC keeps one estimate per bin, so every served prompt
updates the estimate for all prompts in its bin (Appendix~\ref{app:kalman}). For this reason,
we compare against the cluster-level method SPaCe and not against per-prompt methods.

\paragraph{Selection by success rate in other settings.} Work on agents that learn from
their own experience arrives at a similar rule, which is to train on tasks that the model
completes sometimes but not always. Echoverse \citep{pandya2026echoverse} keeps the tasks that
a computer-use agent solves in one to three of four attempts. WebRL \citep{qi2024webrl} keeps
generated web tasks whose critic score, an estimate of the task's value, lies between 0.05 and
0.75. In self-play for reasoning, R-Zero \citep{huang2025rzero} rewards a problem proposer most
when about half of the solver's answers agree with their majority vote, and Absolute Zero
\citep{zhao2025absolutezero} rewards the proposer with one minus the solver's success rate, and with zero when the
solver never succeeds. In curriculum design for deep RL, SFL \citep{rutherford2024sfl} scores
a task by its learnability $p(1-p)$, which is the squared reward spread that TSC tracks. An
older line of work scores tasks by learning progress, the absolute rate at which the
performance on a task changes, so that tasks being forgotten are also revisited
\citep{matiisen2017tscl,portelas2019teacher}. LIMR \citep{li2025limr} keeps the
math prompts whose success over training follows the model's overall learning curve. TSC uses
the current reward spread and not learning progress. The two rules agree when progress is
fastest at intermediate difficulty. They differ on a prompt that the model still fails but is
starting to learn.

\paragraph{How these methods are evaluated.} As far as we can tell, each method above is
reported with a fixed number of prompts per step and one learning rate. Where to Spend Rollouts
\citep{wang2026wheretospend} proposes HORA, which redistributes a fixed rollout budget per
step across a fixed batch of prompts. GVM-RAFT \citep{yao2025gvmraft} chooses the number of rollouts per
prompt to reduce gradient variance. To our knowledge, no prior comparison of selection methods
varies the number of prompts per step against the number of rollouts per prompt, or tunes the
learning rate for each configuration. Section~\ref{sec:adding} shows that both choices can
change which method appears better.

\section{Method}\label{sec:method}

\subsection{Zero-Variance Groups in GRPO}\label{sec:setup}
Each optimizer step uses $P$ prompts. For each prompt, the policy generates $G$ answers, which
we call \emph{rollouts}, so a step generates $P\cdot G$ rollouts. The total number of
rollouts is the rollout budget. An exact-match check of the
final answer gives each rollout a reward $r_i\in\{0,1\}$ (we study non-binary rewards on an
agent task in Section~\ref{sec:agent}). We write a \emph{batch configuration}
as $\shape{P}{G}$. For example, $\shape{8}{8}$ uses eight prompts per step with eight rollouts
each, and $\shape{32}{4}$ uses 32 prompts with four rollouts each. The GRPO advantage of rollout
$i$ is
\begin{equation}
A_i \;=\; \frac{r_i - \bar r}{s},
\qquad \bar r = \tfrac1G\textstyle\sum_i r_i,\quad
s = \text{sample standard deviation of } \{r_i\},
\label{eq:grpo}
\end{equation}
and the advantages enter a clipped surrogate objective with a KL penalty to the reference
model. When all $G$ rewards agree, $s=0$ and every $A_i$ is set to zero. A prompt that the model
solves with probability $p$ produces a zero-variance group with probability $p^G+(1-p)^G$, so
both very hard and very easy prompts waste generation. No sampler can create a gradient on a
prompt with $p=0$. Selection can only move generation to the prompts where a gradient exists.

\subsection{Reward Variance and Advantage Magnitude}\label{sec:identity}
Let $\sigp=\sqrt{k(G-k)}/G$ denote the population standard deviation of the rewards in a group
with $k$ correct rollouts out of $G$. We call $\sigp$ the \emph{reward spread} of the group.

\begin{proposition}[Advantage-mass identity]\label{prop:identity}
Under \eqref{eq:grpo}, for every $k\in\{1,\dots,G-1\}$,
\[
\sum_{i=1}^{G} |A_i| \;=\; 2\sqrt{G(G-1)}\;\sigp .
\]
\end{proposition}

The proof is in Appendix~\ref{app:proofs}. The identity says that the reward spread of a
group equals, up to a constant, the total advantage magnitude that the group contributes to
the update. A sampler that serves groups with a large spread therefore serves groups with a
large update. For binary rewards, the squared spread equals $\mu(1-\mu)$, where $\mu$ is the
mean reward of the group, so the mean and the spread carry the same information.

\subsection{The Thompson-Sampled Curriculum}\label{sec:sampler}
TSC needs an estimate of the reward spread that a prompt will produce before the prompt is
served. Estimating the spread for each prompt separately is not feasible at our budgets, because the
average prompt is served less than once in a run. TSC therefore pools prompts into nine bins
by the untrained model's solve count out of eight, $k\in\{0,\dots,8\}$, which is measured once
before training (for the non-binary rewards of the agent task, we use five bins of equal size
by each question's mean reward; Section~\ref{sec:agent}). Bins defined this way separate prompts by reward spread much better than
clusters of topic and difficulty (Appendix~\ref{app:kalman}).

Figure~\ref{fig:tsc} shows how TSC fits into a GRPO training step, and Algorithm~\ref{alg:tsc}
gives the full procedure. Each bin $c$ has a Gaussian belief $\mathcal N(m_c,v_c)$ over the
reward spread of its prompts. To fill each of the $P$ slots of a step, TSC draws
$\tilde u_c\sim\mathcal N(m_c,v_c)$ for every bin, selects the bin with the largest draw, and
serves a prompt chosen uniformly from that bin. The policy then generates $G$ rollouts for each
selected prompt, the verifier scores them, and GRPO updates the policy as usual. Only the
choice of prompts differs from uniform sampling, and no generated rollout is discarded.

After the rollouts are scored, TSC updates the belief of each served bin by a scalar Kalman
step toward the observed spread $\hat\sigma_p$ of its group. The variance of every bin that
was not served grows by a small amount $\tau$. This growth lets TSC follow a changing model.
The variance growth is a simpler alternative to the change-point detection used for Thompson sampling in
changing environments \citep{mellor2013switching}. A bin that has
not been served for a while becomes uncertain enough to be drawn again, so TSC notices when a
bin that gave no signal earlier starts to give some.

Prior work disagrees on what to do with prompts that currently produce zero-variance groups.
MobileRL \citep{xu2025mobilerl}, an online RL method for agents that operate mobile phones,
lowers the sampling weight of a task after it returns only zero rewards for two consecutive
epochs, and removes the task after a further cooldown of three epochs. A study of RL for search
agents \citep{coelho2026recycling} keeps such queries in training and reports that, in a run with
Qwen3-1.7B, about 20\% of the unique queries change from zero variance to nonzero variance
during training. TSC never removes a bin. It
serves a bin rarely while the bin gives no signal, and the variance growth brings the bin back
if the model's success rate on it changes.

One training step can be written as
\begin{equation}\label{eq:step}
\mathcal S_t\sim q_t,\qquad \theta_{t+1}=U(\theta_t,\mathcal S_t),\qquad q_{t+1}=K(q_t,\hat\sigma_t),
\end{equation}
where $q_t$ is the sampler's distribution over prompts at step $t$, $\mathcal S_t$ is the set
of $P$ prompts it selects, $U$ is generation, scoring and the GRPO update, and $K$ is the
belief update from the observed reward spreads $\hat\sigma_t$. Uniform sampling keeps $q_t$
uniform over the pool and has no $K$. In all of our comparisons, $U$ is the same, including the
model, data, generation length and training recipe. A difference in accuracy between the two
methods therefore comes from the choice of $q_t$ and from the learning rate, which we tune for
each method.

We use Thompson sampling instead of always serving the bin with the highest mean, because a
greedy rule rarely revisits bins whose value is uncertain. The constants are set from measured
noise and drift and not by search (Appendix~\ref{app:kalman}). We use a prior mean $m_0=0.3$,
a prior variance $v_0=0.01$, an observation noise $R=0.0356$ equal to the measured per-group
variance of $\hat\sigma_p$, and a variance growth $\tau=10^{-5}$ applied once per optimizer
step. The Thompson draws and the belief update take microseconds per step. This
latency is negligible, because generating the rollouts of a step and training on them takes
about one minute (Appendix~\ref{app:repro}).

\begin{algorithm}[t]
\caption{Thompson-sampled curriculum (TSC) for GRPO}
\label{alg:tsc}
\begin{algorithmic}[1]
\Require prompt pool $\mathcal D$; solve counts $k(x)\in\{0,\dots,8\}$ of the untrained model for every $x\in\mathcal D$;
prompts per step $P$; rollouts per prompt $G$; steps $T$; prior $(m_0,v_0)$; observation noise $R$;
variance growth $\tau$
\State $\mathcal B_c\gets\{x\in\mathcal D: k(x)=c\}$ for $c=0,\dots,8$ \Comment{bins, built once}
\State $m_c\gets m_0,\; v_c\gets v_0$ for all $c$
\For{$t=1,\dots,T$}
  \State $\mathcal S\gets\emptyset$
  \For{$j=1,\dots,P$} \Comment{select before generating}
    \State draw $\tilde u_c\sim\mathcal N(m_c,v_c)$ for every bin $c$
    \State $c_j\gets\arg\max_c \tilde u_c$;\quad draw $x_j$ uniformly from $\mathcal B_{c_j}$;\quad $\mathcal S\gets\mathcal S\cup\{(x_j,c_j)\}$
  \EndFor
  \For{each $(x_j,c_j)\in\mathcal S$}
    \State generate $y_{j,1},\dots,y_{j,G}\sim\pi_\theta(\cdot\mid x_j)$ and score $r_{j,i}\in\{0,1\}$
  \EndFor
  \State update $\theta$ with the GRPO objective, using advantages \eqref{eq:grpo} computed within each group
  \For{each $(x_j,c_j)\in\mathcal S$} \Comment{Kalman step for the served bin}
    \State $z\gets\hat\sigma_p(r_{j,1},\dots,r_{j,G})$;\quad $v'\gets v_{c_j}+\tau$;\quad $g\gets v'/(v'+R)$
    \State $m_{c_j}\gets m_{c_j}+g\,(z-m_{c_j})$;\quad $v_{c_j}\gets(1-g)\,v'$
  \EndFor
  \For{each bin $c$ not served at step $t$}
    \State $v_c\gets\min(v_c+\tau,\;v_0)$ \Comment{uncertainty grows while unserved}
  \EndFor
\EndFor
\end{algorithmic}
\end{algorithm}

\subsection{Relation to Other Selection Signals}\label{sec:unification}
For binary rewards, each selection signal we know of in this literature is a function of the
solve probability $p$. TSC targets $\sqrt{p(1-p)}$, VCRL's variance is $p(1-p)$, and MoPPS
prefers $p$ close to $\tfrac12$. These three peak at $p=\tfrac12$ and rank prompts in the same
order. They differ only in how quickly they fall off toward $p=0$ and $p=1$. SPaCe's
preference for low reward corresponds to $1-p$, which is largest at $p=0$. On a pool with many
never-solved prompts, this signal points at exactly those prompts, and SPaCe adds a penalty for
clusters that stop improving to move away from them. We test SPaCe in Section~\ref{sec:comparators}.

\section{Experimental Setup}\label{sec:exp}

\paragraph{Models and data.} The main model is Qwen3-1.7B \citep{qwen3} in non-thinking mode,
with a budget of 4{,}096 tokens per answer. We train it on DeepScaleR
\citep{luo2025deepscaler}, a collection of about 40{,}000 competition-style mathematics
problems. Of the 40{,}308 problems in the copy we use, we remove 1{,}531 duplicates and 7 problems that also appear in MATH500, and hold out
300 problems for validation and 500 for testing, which leaves a training pool of 37{,}970
problems. The
reward is 1 when the boxed final answer matches the reference and 0 otherwise. This model has
room to improve on this pool, since its success rate on the prompts that uniform sampling serves
is 39\%. The bins come from one pass of the untrained model over the pool with eight rollouts
per prompt. Section~\ref{sec:second-model} repeats the main configurations on Qwen3-0.6B and
reports a comparison on Qwen3-4B, each with bins measured by that model. We also study two
other tasks, code generation with unit tests (Section~\ref{sec:code}) and a research agent
with graded, non-binary rewards (Section~\ref{sec:agent}).

\paragraph{Batch configurations and learning rates.} We use ten batch configurations
$\shape{P}{G}$ with at most 128 rollouts per step. Six have $P\cdot G\in\{16,32,64\}$ and four
have $P\cdot G=128$. For each configuration and each sampler, we try at least three learning
rates, and we extend the range until both a lower rate and a higher rate score worse than the
best rate. Every run
lasts 250 steps and every setting is run with two seeds. This gives 84 combinations of
configuration, sampler and learning rate, and 168 runs. All runs use LoRA \citep{hu2022lora}
and the same training recipe (Appendix~\ref{app:repro}), so runs at the same learning rate differ only in the sampler.

\paragraph{Evaluation.} We evaluate on 500 DeepScaleR problems that are never used in
training. For transfer, we use MATH500 \citep{hendrycks2021math,lightman2023verify} and the 2024
and 2025 AIME exams, which have 30 problems each. For every problem, we draw 16 rollouts and
count the number $c$ that are correct. We report three measures.
\begin{itemize}\itemsep1pt
\item \textbf{avg@16} is the fraction of correct rollouts, $c/16$, averaged over problems. It
estimates the probability that a single rollout is correct and is our main measure of accuracy.
The untrained 1.7B model scores 37.54 on the held-out set.
\item \textbf{pass@16} is the fraction of problems solved by at least one of the 16 rollouts
\citep{chen2021codex}. It measures \emph{coverage}, that is, whether a correct answer is
within the model's reach.
\item \textbf{pass$^{16}$} is the fraction of problems solved by all 16 rollouts
\citep{yao2024taubench}. It measures \emph{reliability}, that is, whether the model solves a
problem every time.
\end{itemize}
For a problem that the model solves with probability $p$, the three measures are $p$,
$1-(1-p)^{16}$ and $p^{16}$. A gain in avg@16 can therefore come from problems with $p$ near
zero, which pass@16 detects, or from problems with $p$ near one, which pass$^{16}$ detects.
Unless stated otherwise, the \emph{gain} of TSC at a configuration is TSC's avg@16 minus
uniform sampling's avg@16 after 250 steps, each at its own best learning rate.

\section{Results}\label{sec:results}

We first describe uniform sampling alone (Section~\ref{sec:uniform}), then add TSC
(Section~\ref{sec:adding}) and compare the two at equal generation cost
(Section~\ref{sec:frontier}). We then analyze what the gain consists of
(Section~\ref{sec:coverage}) and whether we can predict it (Section~\ref{sec:diversity}). The last
four subsections compare TSC with other selection methods (Section~\ref{sec:comparators}),
other model sizes (Section~\ref{sec:second-model}) and two other tasks
(Sections~\ref{sec:code} and~\ref{sec:agent}).

\subsection{Uniform Sampling Across Batch Configurations}\label{sec:uniform}

\begin{figure}[t]
\centering
\begin{minipage}[t]{.49\linewidth}\centering
\begin{tikzpicture}
\begin{axis}[
  width=\linewidth, height=5.0cm, ybar, bar width=7pt,
  xlabel={solve count of the untrained model (out of 8)}, ylabel={share of pool (\%)},
  ymin=0, ymax=40, xtick={0,...,8}, enlarge x limits=0.07,
  tick label style={font=\small}, label style={font=\small},
  grid=major, grid style={line width=.1pt, draw=gray!18},
]
\addplot[bar shift=0pt, fill=cZero!40, draw=cZero!85] coordinates {(0,34.3) (8,17.1)};\addplot[bar shift=0pt, fill=figCon!40, draw=figCon!85] coordinates {(1,12.3) (2,8.5) (3,6.7) (4,5.5) (5,5.3) (6,5.0) (7,5.3)};
\end{axis}
\end{tikzpicture}
\end{minipage}\hfill
\begin{minipage}[t]{.49\linewidth}\centering
\begin{tikzpicture}
\begin{axis}[
  width=\linewidth, height=5.0cm,
  xlabel={rollouts per prompt $G$ (log scale)}, ylabel={zero-variance groups (\%)},
  xmode=log, log basis x=2, xtick={4,8,16,32}, xticklabels={4,8,16,32},
  xmin=3.4, xmax=38, ymin=20, ymax=70,
  legend pos=north east, legend cell align=left, legend style={font=\scriptsize},
  tick label style={font=\small}, label style={font=\small},
  grid=both, grid style={line width=.1pt, draw=gray!18},
]
\addplot[cUnif, thick, mark=*] coordinates {(4,61.8) (8,47.5) (16,37.1) (32,30.7)};
\addplot[black, thin, dashed] coordinates {(4,65.0) (8,52.3) (16,42.0) (32,33.8)};
\legend{{uniform, measured on 1.7B}, {Beta--Binomial, 4B fit}}
\end{axis}
\end{tikzpicture}
\end{minipage}
\caption{\textbf{Zero-variance groups on the DeepScaleR pool.} Left: distribution of the
pool by the number of correct answers among eight Qwen3-1.7B rollouts; the two end bins (red)
together hold 51.4\% of the pool. Right: fraction of zero-variance groups served by
uniform sampling during training, against the number of rollouts per prompt, with the
Beta--Binomial prediction fitted on the 4B pool.}
\label{fig:dead}
\end{figure}
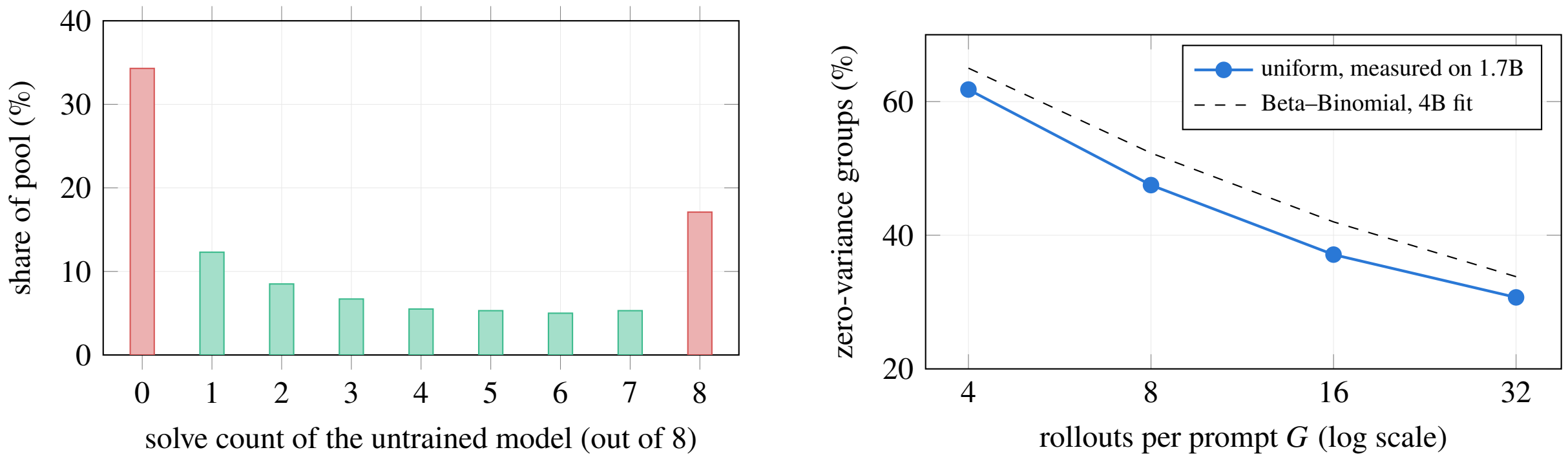

\paragraph{About half of the groups under uniform sampling have zero variance.}
Figure~\ref{fig:dead} (left) shows the pool by the untrained model's solve count. The model
never solves 34.3\% of the prompts and always solves 17.1\%. During training, the share of
zero-variance groups under uniform sampling is 62\%, 48\%, 37\% and 31\% with 4, 8, 16 and 32
rollouts per prompt (Figure~\ref{fig:dead}, right). Each doubling of the rollouts per prompt
removes only about a fifth of the zero-variance groups, because most of them come from
prompts that the model almost never or almost always solves. A Beta--Binomial model of the
pool reproduces this slow decline (Appendix~\ref{app:betabin}).

\paragraph{Uniform sampling does best with few prompts and many rollouts per prompt.} The
best uniform setting is $\shape{4}{16}$ with 41.64\%, and $\shape{4}{32}$ reaches nearly the
same score (Table~\ref{tab:grid}). At 128 rollouts per step, accuracy rises from 40.9\% at
$\shape{32}{4}$ to 41.6\% at $\shape{4}{32}$. More rollouts per prompt mean fewer zero-variance
groups. Adding rollouts per prompt is the only way that uniform sampling can reduce these groups.

\paragraph{The best learning rate depends on the configuration.} The best rate for uniform
sampling ranges from $2.5\times10^{-6}$ to $2\times10^{-5}$ across the ten configurations. At $\shape{4}{4}$, uniform sampling scores 40.95\% at $10^{-5}$ and 36.89\% at
$2\times10^{-5}$. A baseline run at a single learning rate
can therefore be several points below what uniform sampling can reach.

\subsection{Uniform Sampling and TSC}\label{sec:adding}

\paragraph{TSC removes most zero-variance groups in every configuration.} The reduction
ranges from 55\% to 96\% (Table~\ref{tab:grid}). With four rollouts per prompt, the share of
zero-variance groups falls from about 62\% to about 27\%, and at $\shape{4}{32}$, it falls
from 31\% to 1.4\%. TSC achieves this by drawing most of its prompts from the bins with
three to five solved rollouts out of eight (Appendix~\ref{sec:dynamics}). The groups that TSC serves have 1.7 to 2.0 times the reward spread of the
groups that uniform sampling serves. By Proposition~\ref{prop:identity}, they therefore
contribute 1.7 to 2.0 times the total advantage magnitude to the update.

\begin{table}[t]
\centering\footnotesize
\setlength{\tabcolsep}{3.6pt}
\begin{tabular}{lrcrrcrrrrr}
\toprule
 & & \multicolumn{3}{c}{uniform sampling} & \multicolumn{3}{c}{TSC} & & & \\
\cmidrule(lr){3-5}\cmidrule(lr){6-8}
$P{\times}G$ & rollouts/step & rate & avg@16 & pass@16 & rate & avg@16 & pass@16 & gain & removed & entropy\\
\midrule
$\shape{4}{4}$   & 16  & $10^{-5}$ & 40.95 & 75.7 & $10^{-5}$ & 39.59 & 73.4 & $-1.36$ & 55.0\% & 0.97\\
$\shape{8}{4}$   & 32  & $10^{-5}$ & 39.84 & 75.5 & $10^{-5}$ & 40.77 & 75.3 & $+0.93$ & 57.3\% & 1.24\\
$\shape{4}{8}$   & 32  & $10^{-5}$ & 40.70 & 74.9 & $10^{-5}$ & 40.49 & 73.8 & $-0.21$ & 83.2\% & 1.08\\
$\shape{16}{4}$  & 64  & $2{\times}10^{-5}$ & 40.48 & 74.7 & $2{\times}10^{-5}$ & 41.12 & 75.1 & $+0.64$ & 54.9\% & 1.32\\
$\shape{8}{8}$   & 64  & $2.5{\times}10^{-6}$ & 40.24 & 74.5 & $2{\times}10^{-5}$ & 42.25 & 76.6 & $+2.01$ & 76.9\% & 1.40\\
$\shape{4}{16}$  & 64  & $10^{-5}$ & 41.64 & 75.9 & $2{\times}10^{-5}$ & 41.31 & 73.3 & $-0.33$ & 87.9\% & 1.08\\
$\shape{32}{4}$  & 128 & $2{\times}10^{-5}$ & 40.87 & 76.0 & $2{\times}10^{-5}$ & \textbf{42.94} & 76.0 & $+2.07$ & 56.5\% & 1.51\\
$\shape{16}{8}$  & 128 & $2{\times}10^{-5}$ & 41.43 & 74.8 & $2{\times}10^{-5}$ & 42.30 & 75.9 & $+0.87$ & 80.3\% & 1.36\\
$\shape{8}{16}$  & 128 & $10^{-5}$ & 41.45 & 76.1 & $4{\times}10^{-5}$ & 42.25 & 74.4 & $+0.80$ & 84.7\% & 1.46\\
$\shape{4}{32}$  & 128 & $2{\times}10^{-5}$ & 41.59 & 76.5 & $10^{-5}$ & 42.53 & 76.6 & $+0.94$ & 95.5\% & 1.18\\
\midrule
mean & & & 40.92 & 75.5 & & 41.56 & 75.0 & $+0.64$ & & \\
\bottomrule
\end{tabular}
\caption{\textbf{Uniform sampling and TSC on Qwen3-1.7B, by batch configuration}
(held-out set after 250 steps, mean of two seeds, each method at its own best learning
rate; the untrained model scores 37.54 avg@16 and 73.4 pass@16). The gain is TSC's avg@16 minus
uniform sampling's. ``Removed'' is the relative reduction in zero-variance groups under TSC, and
``entropy'' is the entropy in bits of the difficulty bins that TSC serves within one step. Results
by difficulty are in Appendix~\ref{sec:terciles}.}
\label{tab:grid}
\end{table}

\paragraph{The accuracy gain is positive on average and grows with the rollouts per step.}
TSC scores higher than uniform sampling in seven of the ten configurations, and the mean gain
is $+0.64$ points, which is 2.5 standard errors from zero. Grouped by the number of rollouts
per step, the mean gain is $-1.36$ at 16 rollouts, $+0.36$ at 32, $+0.77$ at 64 and $+1.17$ at
128. Over 250 steps, these are budgets of 4{,}000 to 32{,}000 generated rollouts, and
Figure~\ref{fig:headline} (left) shows the same trend for the best accuracy that each method
reaches within each budget. At 128 rollouts per step, TSC is ahead in all four configurations.
The three configurations where TSC is behind all use four prompts per step.

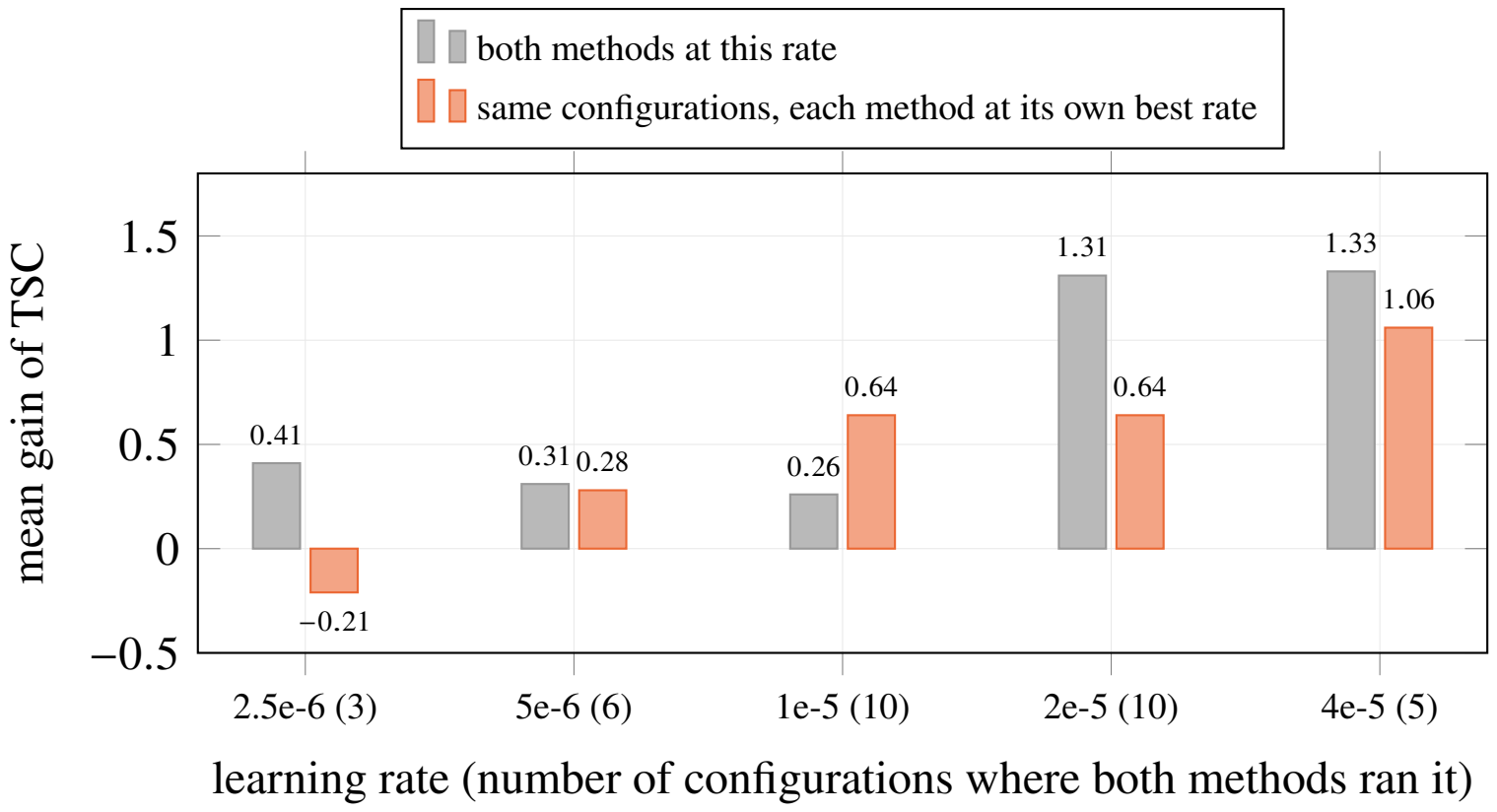
\begin{figure}[t]
\centering
\begin{tikzpicture}
\begin{axis}[
  width=.62\linewidth, height=4.8cm, ybar, bar width=9pt,
  xlabel={learning rate (number of configurations where both methods ran it)},
  ylabel={mean gain of TSC},
  symbolic x coords={2.5e-6 (3),5e-6 (6),1e-5 (10),2e-5 (10),4e-5 (5)}, xtick=data,
  ymin=-0.5, ymax=1.8, ytick={-0.5,0,0.5,1,1.5},
  legend cell align=left, legend style={font=\scriptsize, at={(0.5,1.05)}, anchor=south},
  tick label style={font=\small}, label style={font=\small},
  x tick label style={font=\scriptsize},
  grid=major, grid style={line width=.1pt, draw=gray!18},
  nodes near coords, nodes near coords style={font=\tiny},
  every node near coord/.append style={/pgf/number format/fixed, /pgf/number format/precision=2},
]
\addplot[fill=gray!55, draw=gray!80] coordinates {(2.5e-6 (3),0.41) (5e-6 (6),0.31) (1e-5 (10),0.26) (2e-5 (10),1.31) (4e-5 (5),1.33)};
\addplot[fill=cOurs!60, draw=cOurs] coordinates {(2.5e-6 (3),-0.21) (5e-6 (6),0.28) (1e-5 (10),0.64) (2e-5 (10),0.64) (4e-5 (5),1.06)};
\legend{both methods at this rate, {same configurations, each method at its own best rate}}
\end{axis}
\end{tikzpicture}
\caption{\textbf{Mean gain of TSC under a shared learning rate.} Each pair of bars covers
the configurations where both methods were run at the learning rate on the horizontal axis.
The number of these configurations is given in parentheses. Gray: the mean gain when both
methods use that rate. Orange: the mean gain over the same configurations when each method
uses its own best rate. The orange bars do not use the rate on the axis, and they differ
across the axis only because the set of configurations differs.}
\label{fig:sharedlr}
\end{figure}

\paragraph{The two methods prefer different learning rates.} The best learning rates of the
two methods differ in four of the ten configurations. At $\shape{8}{8}$, the best rate for TSC
is eight times the best rate for uniform sampling, and at $\shape{4}{32}$, the best rate for TSC is half the best rate for uniform sampling. A
comparison at one shared rate therefore favors whichever method that rate suits
(Figure~\ref{fig:sharedlr}). If both methods use $10^{-5}$, the mean gain is $+0.26$, and if
both use $2\times10^{-5}$, it is $+1.31$. With each method at its own best rate, the mean gain
is $+0.64$. Appendix~\ref{app:grid} shows the accuracy at every learning rate.

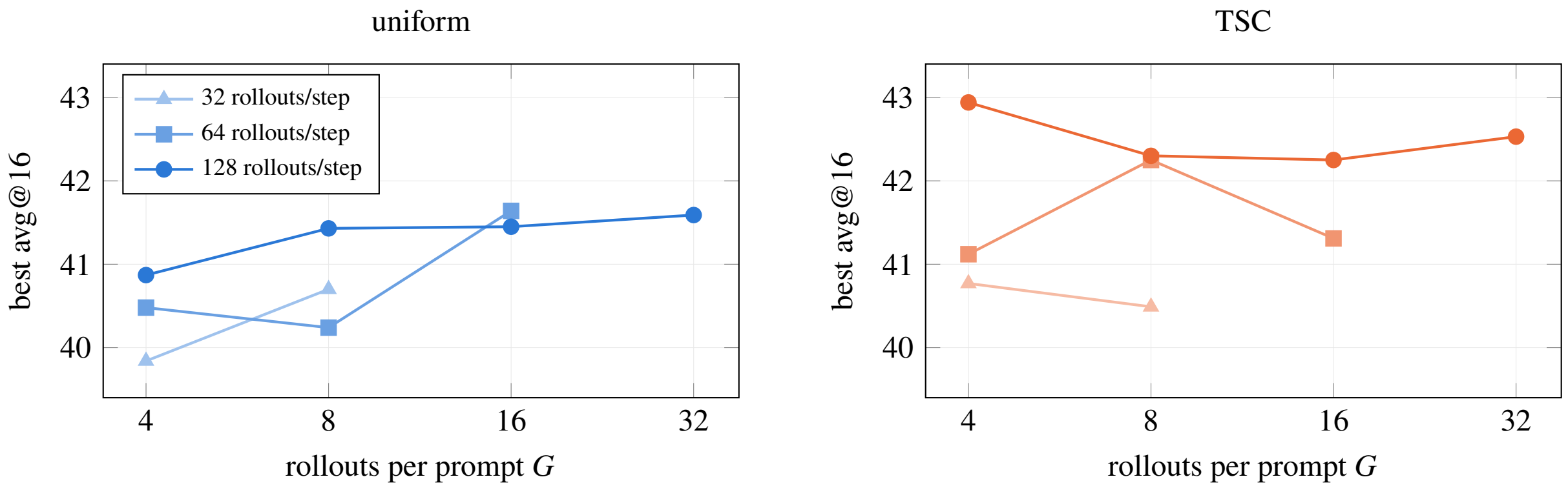
\begin{figure}[t]
\centering
\begin{minipage}[t]{.49\linewidth}\centering
\begin{tikzpicture}
\begin{axis}[
  width=\linewidth, height=5.0cm, title={uniform}, title style={font=\small},
  xlabel={rollouts per prompt $G$}, ylabel={best avg@16},
  xmode=log, log basis x=2, xtick={4,8,16,32}, xticklabels={4,8,16,32},
  xmin=3.4, xmax=38, ymin=39.4, ymax=43.4,
  legend pos=north west, legend cell align=left, legend style={font=\scriptsize},
  tick label style={font=\small}, label style={font=\small},
  grid=both, grid style={line width=.1pt, draw=gray!18},
]
\addplot[cUnif!45, thick, mark=triangle*] coordinates {(4,39.84) (8,40.70)};
\addplot[cUnif!70, thick, mark=square*] coordinates {(4,40.48) (8,40.24) (16,41.64)};
\addplot[cUnif, thick, mark=*] coordinates {(4,40.87) (8,41.43) (16,41.45) (32,41.59)};
\legend{32 rollouts/step, 64 rollouts/step, 128 rollouts/step}
\end{axis}
\end{tikzpicture}
\end{minipage}\hfill
\begin{minipage}[t]{.49\linewidth}\centering
\begin{tikzpicture}
\begin{axis}[
  width=\linewidth, height=5.0cm, title={TSC}, title style={font=\small},
  xlabel={rollouts per prompt $G$}, ylabel={best avg@16},
  xmode=log, log basis x=2, xtick={4,8,16,32}, xticklabels={4,8,16,32},
  xmin=3.4, xmax=38, ymin=39.4, ymax=43.4,
  legend pos=south east, legend cell align=left, legend style={font=\scriptsize},
  tick label style={font=\small}, label style={font=\small},
  grid=both, grid style={line width=.1pt, draw=gray!18},
]
\addplot[cOurs!45, thick, mark=triangle*] coordinates {(4,40.77) (8,40.49)};
\addplot[cOurs!70, thick, mark=square*] coordinates {(4,41.12) (8,42.25) (16,41.31)};
\addplot[cOurs, thick, mark=*] coordinates {(4,42.94) (8,42.30) (16,42.25) (32,42.53)};
\end{axis}
\end{tikzpicture}
\end{minipage}
\caption{\textbf{Best accuracy by rollouts per prompt at a fixed number of rollouts per
step.} Each line holds the number of rollouts per step fixed, so moving right means fewer
prompts per step with more rollouts each. Uniform sampling (left) does best with many rollouts
per prompt. TSC (right) does best with four rollouts per prompt at 128 rollouts per step.}
\label{fig:configurations}
\end{figure}

\paragraph{The two methods prefer opposite batch configurations.}
Figure~\ref{fig:configurations} holds the number of rollouts per step fixed and varies how it
is divided. Uniform sampling does best with four prompts and many rollouts per prompt. TSC
does best at $\shape{32}{4}$, with many prompts and four rollouts each. TSC already chooses prompts whose rewards vary, so it does not
need many rollouts per prompt to obtain a mixed group. It benefits instead from more prompts
per step, a point we return to in Section~\ref{sec:diversity}. A comparison at a single
configuration can therefore favor either method. At $\shape{4}{16}$, uniform sampling is ahead
by 0.33 points, and at $\shape{32}{4}$, TSC is ahead by 2.07. This result shows why a
selection method should be compared with uniform sampling at several batch configurations
before a gain is reported. As far as we can tell, the selection methods in
Section~\ref{sec:related} are each reported at one batch configuration.

\paragraph{The gain transfers to other benchmarks.} We scored the two configurations with the largest gain, $\shape{8}{8}$ and $\shape{32}{4}$, on
MATH500 and on the two AIME exams. On MATH500, TSC is ahead of uniform sampling by 0.9 points
at $\shape{8}{8}$ and by 1.0 at $\shape{32}{4}$, about half of the gain on the held-out set.
The AIME results point in the same direction but have wide intervals, because each exam has
only 30 problems (Appendix~\ref{sec:transfer}).

\subsection{Accuracy and Generation Cost}\label{sec:frontier}

Both methods generate the same number of rollouts in a given configuration, so we can compare
them at equal cost. Figure~\ref{fig:headline} (left) shows the best accuracy that each method
reaches with a given total number of generated rollouts or fewer, from 4{,}000 to 32{,}000.

\paragraph{TSC raises the highest accuracy that can be reached.} Uniform sampling levels off
at 41.6\% from 16{,}000 rollouts on. TSC reaches 42.3\% with 16{,}000 rollouts and 42.9\% with
32{,}000. Only TSC reaches 42\% within the budgets we tested.

\paragraph{Uniform sampling is cheaper for lower targets.} Uniform sampling reaches 40.0\%
and 40.5\% with 4{,}000 rollouts, and TSC needs 8{,}000. Both methods need the same budget to
reach 41.0\% and 41.5\%. TSC is therefore not a cheaper route to a modest target. It pays off
when the goal is the best model that a larger budget can produce.

\paragraph{For high accuracy, TSC is the more efficient method.}
Figure~\ref{fig:headline} (right) plots every configuration by its accuracy and by its
accuracy per 1{,}000 generated rollouts. No uniform setting scores above 41.64\%. TSC at
$\shape{8}{8}$ scores 42.25\% with 16{,}000 rollouts. This is above every uniform setting,
including the four that generate 32{,}000 rollouts. The mean gain over the four configurations with 32{,}000 rollouts is $+1.17$ points,
which is 2.9 standard errors from zero.

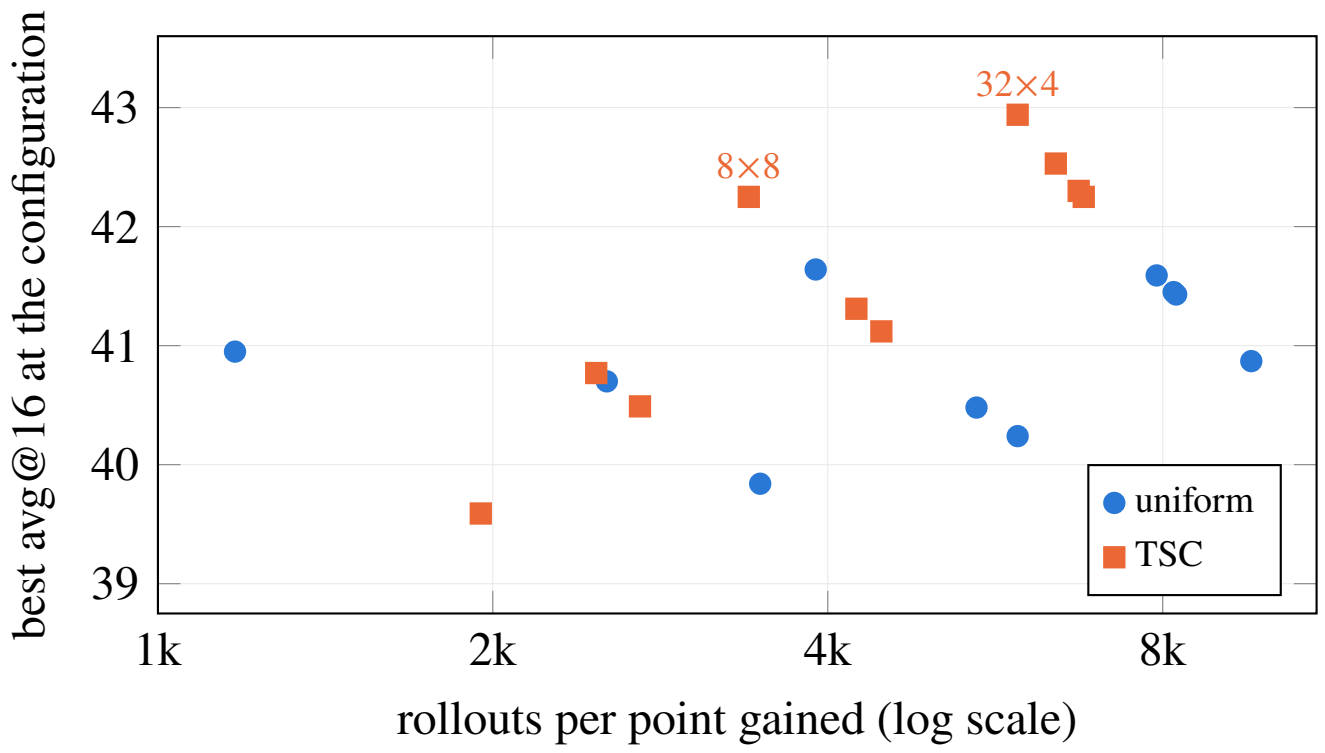
\begin{figure}[t]
\centering
\begin{tikzpicture}
\begin{axis}[
  width=.56\linewidth, height=5.4cm,
  xlabel={rollouts per point gained (log scale)},
  ylabel={best avg@16 at the configuration},
  xmode=log, log basis x=10,
  xmin=1000, xmax=11000, ymin=38.75, ymax=43.6, ytick={39,40,41,42,43},
  xtick={1000,2000,4000,8000}, xticklabels={1k,2k,4k,8k},
  legend pos=south east, legend cell align=left, legend style={font=\scriptsize},
  tick label style={font=\small}, label style={font=\small},
  grid=both, grid style={line width=.1pt, draw=gray!18},
]
\addplot[only marks, mark=*, cUnif, mark size=1.9pt] coordinates {
  (1173,40.95) (3478,39.84) (2532,40.70) (5442,40.48) (5926,40.24)
  (3902,41.64) (9610,40.87) (8226,41.43) (8184,41.45) (7901,41.59)};
\addplot[only marks, mark=square*, cOurs, mark size=1.9pt] coordinates {
  (1951,39.59) (2477,40.77) (2712,40.49) (4469,41.12) (3397,42.25)
  (4244,41.31) (5926,42.94) (6723,42.30) (6794,42.25) (6413,42.53)};
\node[font=\scriptsize, cOurs, anchor=south] at (axis cs:3397,42.25) {8$\times$8};
\node[font=\scriptsize, cOurs, anchor=south] at (axis cs:5926,42.94) {32$\times$4};
\legend{uniform, TSC}
\end{axis}
\end{tikzpicture}
\caption{\textbf{Generation cost per point of accuracy (Qwen3-1.7B, 250 steps, two
seeds).} One point per configuration, plotting best avg@16 against the number of rollouts
generated per point of avg@16 gained over the untrained model. Points further to the upper
left are better.}
\label{fig:frontier}
\end{figure}

\paragraph{Cost per point of accuracy.} Figure~\ref{fig:frontier} divides the rollouts
generated by the points of avg@16 gained over the untrained model. In the seven
configurations where TSC is ahead, uniform sampling needs 1.2 to 1.7 times as many rollouts per point as
TSC. In the three configurations with four prompts per step where TSC is behind,
TSC needs more rollouts per point. Overall, TSC Pareto-dominates uniform sampling in the high-accuracy regime.
TSC at $\shape{8}{8}$ has a higher accuracy and a lower cost per point than each of the seven
uniform settings that generate 16{,}000 or 32{,}000 rollouts.

\paragraph{The gain remains with longer training.} We continued both methods to 1{,}000
steps at $\shape{8}{8}$ and $\shape{32}{4}$. Averaged over the checkpoints from step 600 to
step 1{,}000, TSC scores 41.85\% against 40.52\% at $\shape{8}{8}$ and 42.23\% against 41.16\%
at $\shape{32}{4}$. In both configurations, each TSC run is above each uniform run. Uniform
sampling still serves about 48\% zero-variance groups at $\shape{8}{8}$ throughout, so longer
training does not remove the problem that TSC addresses (Appendix~\ref{app:durability}).

\subsection{Reliability and Coverage}\label{sec:coverage}

A higher avg@16 can mean that the model solves problems it could not solve before, or that
it solves the same problems more often. We separate the two with pass@$k$ and pass$^k$,
computed for every $k\le16$ from the number $c$ of correct rollouts among 16. A problem
contributes $1-\binom{16-c}{k}/\binom{16}{k}$ to pass@$k$ and $\binom{c}{k}/\binom{16}{k}$ to
pass$^k$.

\begin{table}[t]
\centering\small
\setlength{\tabcolsep}{4.5pt}
\begin{tabular}{lcccccc}
\toprule
 & avg@16 & pass@16 & pass$^{16}$ & first solved & no longer solved & newly solved 16/16\\
\midrule
untrained model & 37.54 & 73.4 & 12.8 & --- & --- & ---\\
uniform sampling & 40.92 & 75.5 & 14.2 & 36.5 & 26.2 & 16.4\\
TSC & 41.56 & 75.0 & 15.1 & 36.1 & 27.9 & 21.3\\
\midrule
TSC $-$ uniform & $+0.64$ & $-0.42$ & $+0.89$ & $-0.4$ & $+1.7$ & $+4.9$\\
\bottomrule
\end{tabular}
\caption{\textbf{Reliability and coverage on Qwen3-1.7B} (held-out set of 500 problems, 250
steps, mean over the ten configurations of two seeds at each method's best learning rate).
The last three columns count problems per run, relative to the untrained model. ``First
solved'' problems are solved at least once when the untrained model never solved them. ``No
longer solved'' problems are never solved when the untrained model solved them at least once.
``Newly solved 16/16'' problems are solved in all 16 rollouts when the untrained model did not
do so.}
\label{tab:coverage}
\end{table}

\begin{figure}[t]
\centering
\begin{tikzpicture}
\begin{axis}[
  width=.62\linewidth, height=5.0cm,
  xmode=log, log basis x=2, xmin=0.85, xmax=19,
  xtick={1,2,4,8,16}, xticklabels={1,2,4,8,16},
  xlabel={number of rollouts $k$}, ylabel={gain of TSC over uniform},
  ymin=-0.7, ymax=1.3,
  tick label style={font=\small}, label style={font=\small},
  legend pos=south west, legend cell align=left, legend style={font=\scriptsize},
  grid=major, grid style={line width=.1pt, draw=gray!16},
]
\addplot[gray, thin, forget plot] coordinates {(0.85,0) (19,0)};
\addplot[cOurs, thick, mark=square*, mark size=2pt] coordinates {(1,0.64) (2,0.85) (4,1.00) (8,1.05) (16,0.89)};
\addplot[cUnif, thick, mark=*, mark size=2pt] coordinates {(1,0.64) (2,0.43) (4,0.22) (8,-0.08) (16,-0.42)};
\legend{{pass$^k$ (all $k$ correct)}, {pass@$k$ (at least one correct)}}
\end{axis}
\end{tikzpicture}
\caption{\textbf{Gain of TSC over uniform sampling as a function of $k$} (Qwen3-1.7B,
mean over the ten configurations of two seeds at each method's best learning rate). At $k=1$,
both measures equal avg@16.}
\label{fig:coverage}
\end{figure}
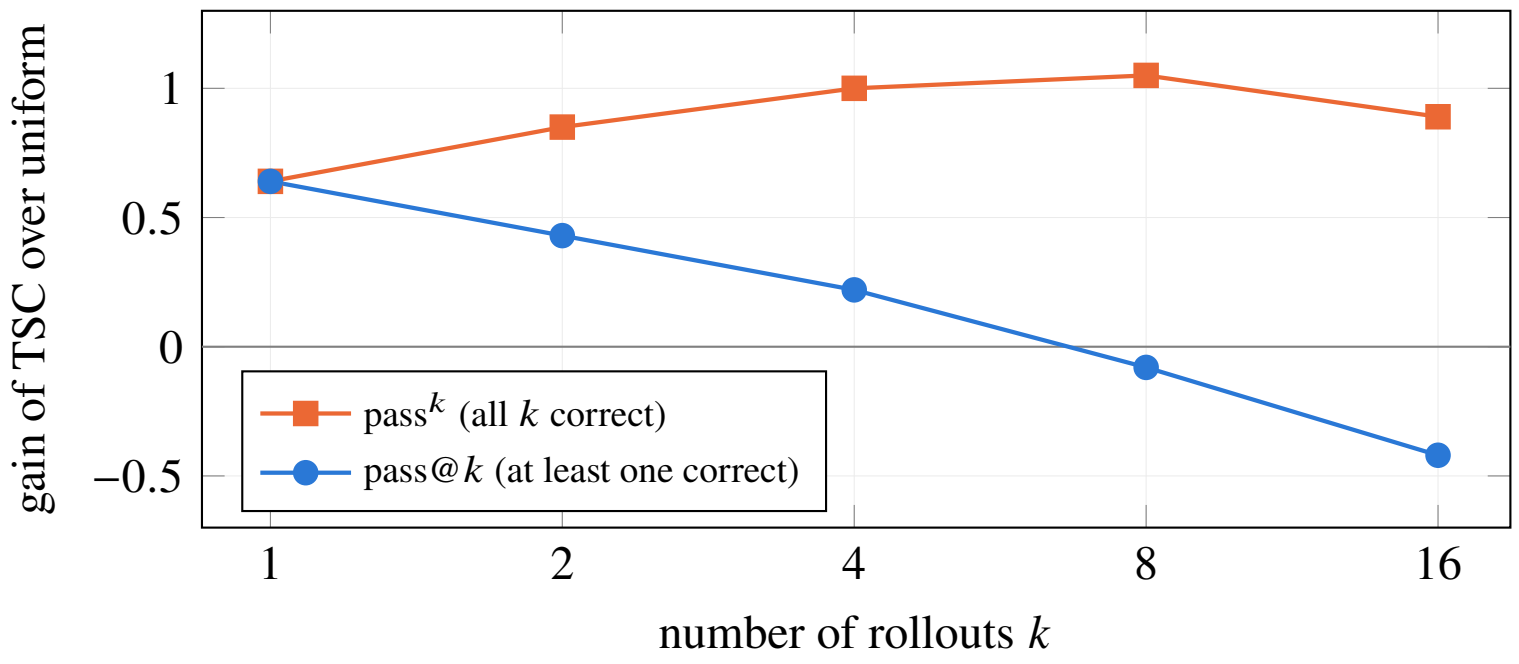

\paragraph{The gain is in reliability, not coverage.} Table~\ref{tab:coverage} averages the
ten configurations. TSC reaches a pass$^{16}$ of 15.1\% and uniform sampling reaches 14.2\%, against 12.8\% for the
untrained model. The difference is positive in nine of the ten configurations. TSC's
pass@16 is 75.0\% against 75.5\% for uniform sampling, a difference within noise.
Figure~\ref{fig:coverage} shows the same pattern for every $k$. The gain in pass$^k$ grows
from $+0.64$ at $k=1$ to about $+1.0$ for $k\ge4$, and the gain in pass@$k$ shrinks to
$-0.42$ at $k=16$. TSC trains mostly on prompts that the model solves some of the time, and
the result is a model that solves such problems more consistently.

\paragraph{The problems that become reliable are of medium difficulty.} A TSC run makes
21.3 problems newly reliable, that is, solved in all 16 rollouts when the untrained model did
not do so. A uniform run makes 16.4 problems newly reliable. Most of these problems were
already solved 12 to 15 times out of 16 by the untrained model. When we divide the held-out
set into three equal groups by the untrained model's accuracy, the gain in avg@16 is $+0.28$
on the easy third, $+1.02$ on the medium third and $+0.61$ on the hard third
(Appendix~\ref{sec:terciles}).

\paragraph{Neither method expands coverage by much.} A model trained with either method solves
about 36 problems that the untrained model never solved. Two seeds of the same setting differ
by about 32 such problems, so most of these changes come from sampling and not from training.
This agrees with the observation that RL with verifiable rewards mainly sharpens abilities
that the untrained model already has \citep{yue2025rlbeyond}.

\subsection{Predictors of the Gain}\label{sec:diversity}

The gain of TSC ranges from $-1.4$ to $+2.1$ points across the ten configurations. We now analyze what determines whether TSC helps at a given configuration.

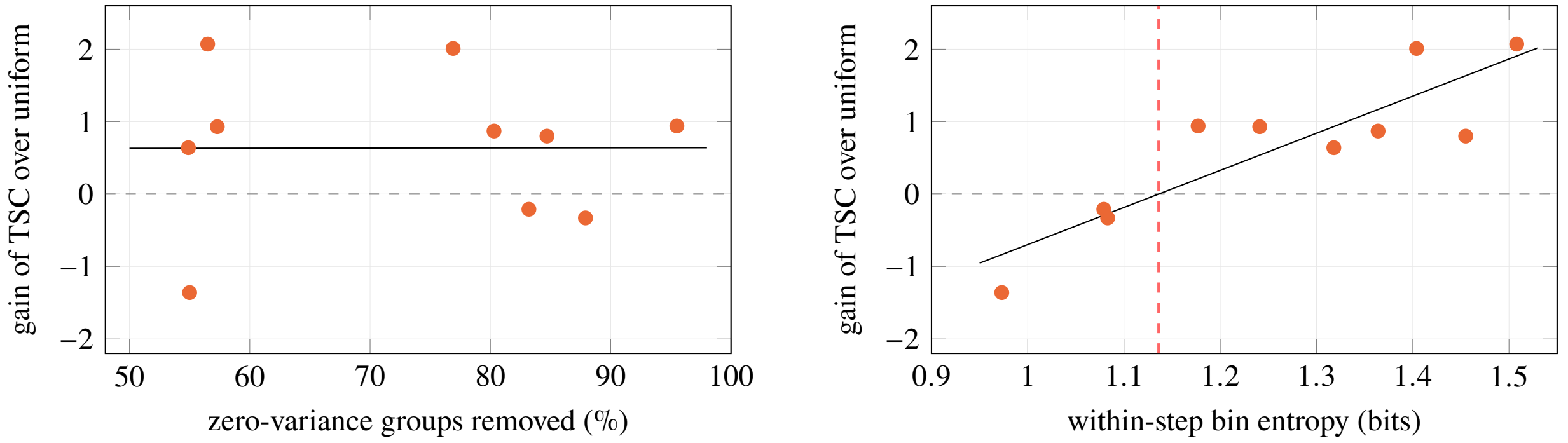
\begin{figure}[t]
\centering
\begin{minipage}[t]{.49\linewidth}\centering
\begin{tikzpicture}
\begin{axis}[
  width=\linewidth, height=5.2cm,
  xlabel={zero-variance groups removed (\%)},
  ylabel={gain of TSC over uniform},
  xmin=48, xmax=100, ymin=-2.2, ymax=2.6, ytick={-2,-1,0,1,2},
  tick label style={font=\small}, label style={font=\small},
  grid=both, grid style={line width=.1pt, draw=gray!18},
]
\addplot[gray, dashed, thin] coordinates {(48,0) (100,0)};
\addplot[only marks, mark=*, cOurs, mark size=2.0pt] coordinates {
  (55.0,-1.36) (57.3,0.93) (83.2,-0.21) (54.9,0.64) (76.9,2.01)
  (87.9,-0.33) (56.5,2.07) (80.3,0.87) (84.7,0.80) (95.5,0.94)};
\addplot[black, thin] coordinates {(50,0.632) (98,0.640)};
\end{axis}
\end{tikzpicture}
\end{minipage}\hfill
\begin{minipage}[t]{.49\linewidth}\centering
\begin{tikzpicture}
\begin{axis}[
  width=\linewidth, height=5.2cm,
  xlabel={within-step bin entropy (bits)},
  ylabel={gain of TSC over uniform},
  xmin=0.9, xmax=1.55, ymin=-2.2, ymax=2.6, ytick={-2,-1,0,1,2},
  tick label style={font=\small}, label style={font=\small},
  grid=both, grid style={line width=.1pt, draw=gray!18},
]
\addplot[gray, dashed, thin] coordinates {(0.9,0) (1.55,0)};
\addplot[red!60, thick, dashed] coordinates {(1.136,-2.2) (1.136,2.6)};
\addplot[black, thin] coordinates {(0.95,-0.953) (1.53,2.018)};
\addplot[only marks, mark=*, cOurs, mark size=2.0pt] coordinates {
  (0.973,-1.36) (1.241,0.93) (1.079,-0.21) (1.318,0.64) (1.404,2.01)
  (1.083,-0.33) (1.508,2.07) (1.364,0.87) (1.455,0.80) (1.177,0.94)};
\end{axis}
\end{tikzpicture}
\end{minipage}
\caption{\textbf{Reduction in zero-variance groups and within-step entropy against the
accuracy gain of TSC over uniform sampling.} Each point is one of the ten configurations (1.7B, 250 steps, two
seeds). Left: relative reduction in zero-variance groups under TSC; the fitted line is flat
($r\approx0.00$). Right: entropy of the difficulty bins served within one step, over the first
250 steps; the fitted line has a slope of 5.1 points per bit ($R^2=0.77$) and crosses zero at
1.14 bits (red line).}
\label{fig:mechanism}
\end{figure}

\paragraph{The share of zero-variance groups removed does not predict the gain.} Curricula
of this kind are motivated by the idea that removing zero-variance groups makes each
update more informative, and that accuracy then improves. TSC does remove zero-variance
groups in every configuration, but the share it removes is unrelated to the gain
(Figure~\ref{fig:mechanism}, left). The correlation across the ten configurations is
$r\approx0.00$, with a 95\% interval of $[-0.63,+0.63]$. At $\shape{4}{32}$, TSC removes the
largest share (95.5\%) and gains $+0.9$ points. At $\shape{32}{4}$, it removes 56.5\% and gains
the most ($+2.1$). At $\shape{4}{4}$, it removes a similar 55\% and falls behind by 1.4 points.

\paragraph{The variety of difficulty within a step predicts the gain.} We measure this
variety as the entropy, in bits, of the bins that TSC serves in one step. A step whose
prompts all come from one or two bins has low entropy, and a step spread over several bins
has high entropy. Across the ten configurations, the gain rises by about 5.1 points per bit
of entropy ($r=0.88$, 95\% interval $[0.55,0.97]$) and crosses zero near 1.14 bits
(Figure~\ref{fig:mechanism}, right). We can therefore estimate
whether TSC helps at a configuration by comparing its within-step entropy with this threshold
of 1.14 bits. In all ten configurations, TSC is ahead when the entropy is above the threshold
and behind when it is below.

\paragraph{Entropy carries more information than the number of prompts.} The entropy of a step with $P$
prompts cannot exceed $\log_2\min(P,9)$ bits, so entropy and $P$ are related. When both are
used as predictors, however, entropy keeps its slope and $\log_2P$ adds nothing
(Appendix~\ref{app:repro}). The one configuration with four prompts where TSC gains,
$\shape{4}{32}$ at 1.18 bits, is predicted correctly by the entropy threshold and incorrectly by a
rule based on $P$. The number of rollouts per prompt does not predict the gain. With eight
prompts per step, the gain is $+0.93$, $+2.01$ and $+0.80$ for 4, 8 and 16 rollouts per prompt.

\paragraph{Entropy is a predictor, and we have not shown that it is the cause.} TSC
concentrates each step on a few bins of medium difficulty, and a step with more prompts can
cover more of these bins. To test whether entropy causes the gain, we raised it directly
inside a fixed configuration, once with a temperature on the Thompson draw and once with a
stratified draw that spreads the prompts of a step over the bins. Entropy rose as intended in
both cases, but accuracy improved by much less than the relationship across configurations
predicts (Appendix~\ref{app:causal}).

\subsection{Comparison with Other Selection Methods}\label{sec:comparators}

We compare TSC with four other ways of choosing prompts, on the 1.7B model at
$\shape{8}{8}$. This configuration gives one of the two largest gains of TSC over uniform
sampling, and a run costs half as much as at $\shape{32}{4}$. As in the main comparison, we
sweep the learning rate of each method until both a lower rate and a higher rate score worse than
the best rate, and we run a second seed at the best rate. The four methods are the following.
\begin{itemize}\itemsep1pt
\item \textbf{Offline filtering} drops the prompts that the untrained model solves zero or
eight times out of eight, a symmetric version of the filter in POLARIS \citep{an2025polaris},
which drops only the solved prompts. It draws prompts uniformly from the remaining 18{,}454.
\item \textbf{Static headroom} uses the same bins as TSC with fixed weights. The weight of a
bin is proportional to its size times $1[k\ge1]-k/8$, so never-solved and always-solved
prompts are not served and harder bins receive more weight.
\item \textbf{DAPO} \citep{yu2025dapo} generates groups for uniformly sampled prompts, drops
the zero-variance groups and generates more until the batch is full. We count every rollout
it generates.
\item \textbf{SPaCe} \citep{do2025sparft} is the published cluster-level curriculum closest
to TSC. It forms clusters by K-means on a sentence embedding joined with a difficulty score.
We instead run its scheduler on 21 clusters that SPaCe and TSC share, formed from the seven
topics of the dataset and three difficulty levels measured with the 1.7B model, so that the
two methods differ only in the selection rule (the last two rows of
Table~\ref{tab:comparators}).
\end{itemize}
Appendix~\ref{app:comparators} describes each implementation, and
Table~\ref{tab:comparators} gives the results.

\begin{table}[t]
\centering\footnotesize\setlength{\tabcolsep}{3.5pt}
\begin{tabular}{llccccc}
\toprule
bins & method & avg@16 & pass@16 & zero-var. & hard/med./easy (\%) & rollouts/step\\
\midrule
9 difficulty bins & uniform & 40.24 & 74.5 & 46\% & 34 / 34 / 32 & 64\\
                  & offline filtering ($k=1$--7) & 41.22 & 74.1 & 18\% & \phantom{0}0 / 68 / 32 & 64\\
                  & static headroom & 41.57 & 75.4 & 19\% & \phantom{0}0 / 86 / 14 & 64\\
                  & TSC (ours) & 42.25 & 76.6 & 11\% & \phantom{0}0 / 78 / 22 & 64\\
                  & DAPO & 42.51 & 75.1 & 0\% (50\% dropped) & 34 / 33 / 33 & 137\\
\midrule
21 topic$\times$difficulty & TSC (ours) & 41.01 & 75.6 & 23\% & \phantom{0}3 / 88 / \phantom{0}9 & 64\\
                  & SPaCe scheduler & 40.65 & 75.3 & 54\% & 54 / 31 / 15 & 64\\
\bottomrule
\end{tabular}
\caption{\textbf{Selection methods at the $8{\times}8$ configuration on Qwen3-1.7B} (250
steps, held-out avg@16, each method at its own best learning rate, mean of two seeds). The
zero-variance column gives the share of groups used for training that have zero variance. For
DAPO, which uses no such groups, the value in parentheses is the share of generated groups it
drops, and rollouts per step counts every generated rollout. The hard/med./easy column gives
the percentage of served prompts that the untrained model solves zero, one to four, and five to eight
times out of eight. For DAPO, it counts every generated prompt.}
\label{tab:comparators}
\end{table}

\paragraph{Static selection recovers part of the gain.} Both static methods score between
uniform sampling and TSC. Offline filtering reaches 41.22\%, which is 0.98 points above
uniform sampling and 1.03 below TSC. Static headroom reaches 41.57\%, which is 1.33 points
above uniform sampling and 0.68 below TSC. Both remove most of the zero-variance groups, with
18\% and 19\% remaining against 11\% for TSC, and their gains over uniform sampling are between half and two thirds of
TSC's gain.

\paragraph{DAPO reaches the same accuracy with about twice as many rollouts.} DAPO scores
42.51\% at its best learning rate, against 42.25\% for TSC. DAPO drops about half of the
groups it generates. To fill its batches, it generates 34{,}280 rollouts in 250 steps, 2.14 times
the 16{,}000 that TSC generates. At TSC's learning rate, DAPO scores 42.26\% and
generates 2.10 times as many rollouts. The 4B model gives the same result. After 500 steps at
$\shape{8}{8}$, the two methods have trained on the same 32{,}000 rollouts. The paired difference, TSC minus
DAPO, is $+0.14$ points \ci{-1.03}{+1.30}, and DAPO has generated 2.28 times as many rollouts
(Appendix~\ref{app:4b}).

\paragraph{SPaCe and TSC reach similar accuracy on the same clusters.} On the 21 clusters,
SPaCe scores 40.65\% and TSC scores 41.01\%. The two methods serve
very different prompts. SPaCe prefers clusters with low reward, so 54\% of its prompts are
ones the untrained model never solves, and 54\% of its groups have zero variance, against 23\% for
TSC. With the nine solve-count bins, TSC scores 42.25\% at this configuration, against 41.01\%
with the 21 clusters. For this dataset, the solve-count bins therefore appear sufficient, and
binning by topic does not appear necessary. This may differ for other datasets.

\paragraph{Fewer zero-variance groups do not imply higher accuracy.} Across the methods in
Table~\ref{tab:comparators}, the share of zero-variance groups does not order the methods by
accuracy. DAPO trains on none and matches TSC. SPaCe trains on 54\% and matches TSC on the
same clusters. The static methods train on fewer than a fifth and recover part of the gain.
This agrees with Section~\ref{sec:diversity}, where the share removed also failed to predict
the gain across configurations. Among the methods that reach the highest accuracy at this
configuration, TSC generates the fewest rollouts.

\subsection{Other Model Sizes}\label{sec:second-model}

\paragraph{The gain repeats on a smaller model.} We repeated three configurations on
Qwen3-0.6B, with bins measured by this model and two seeds per setting. We chose
$\shape{8}{8}$ and $\shape{32}{4}$, the two best configurations for TSC on 1.7B, and
$\shape{4}{4}$, where TSC is behind on 1.7B. TSC is ahead in all three
(Table~\ref{tab:06b}). The gain is $+1.40$ at $\shape{8}{8}$ and $+2.90$ at $\shape{32}{4}$,
the largest gain on either model. At $\shape{4}{4}$, the gain is $+0.90$, which is within
noise. For the 0.6B model, more prompts of the pool give zero-variance groups at eight rollouts (62\%
against 51\%), and all of the increase comes from prompts that the model never solves. At
$\shape{8}{8}$, TSC lowers the share of zero-variance groups from 60\% to 12\%. As on 1.7B,
uniform sampling does not improve when the number of prompts per step increases from $\shape{8}{8}$ to
$\shape{32}{4}$ (25.72\% and 25.08\%), and TSC does (27.12\% and 27.98\%).

\begin{table}[t]
\centering\small
\begin{tabular}{lccccc}
\toprule
$P{\times}G$ & uniform (rate) & TSC (rate) & gain & 1.7B gain & predicted by entropy\\
\midrule
$\shape{8}{8}$ & 25.72 ($10^{-5}$) & 27.12 ($10^{-5}$) & $+1.40$ & $+2.01$ & $+0.72$ \\
$\shape{4}{4}$ & 24.66 ($5{\times}10^{-6}$) & 25.56 ($10^{-5}$) & $+0.90$ & $-1.36$ & $+0.15$ \\
$\shape{32}{4}$ & 25.08 ($10^{-5}$) & 27.98 ($10^{-5}$) & $+2.90$ & $+2.07$ & $+1.85$ \\
\bottomrule
\end{tabular}
\caption{\textbf{Results on Qwen3-0.6B} (held-out avg@16, untrained score 21.50, two seeds
per method at the listed learning rate, 250 steps; for TSC, the second seed at $\shape{8}{8}$ and both seeds at $\shape{32}{4}$ were run at $10^{-5}$ only). The last column gives the difference
predicted by the entropy relationship fitted on 1.7B, at the within-step entropy measured
for TSC.}
\label{tab:06b}
\end{table}

\paragraph{The entropy relationship carries over.} The line fitted on 1.7B predicts the
correct sign of the gain in all three configurations on 0.6B. The measured gains increase in
the same order as TSC's within-step entropy, which is 1.16, 1.28 and 1.50 bits at
$\shape{4}{4}$, $\shape{8}{8}$ and $\shape{32}{4}$, respectively.

\paragraph{The gain is smaller on a larger model.} On Qwen3-4B, we ran one matched pair at
$\shape{8}{8}$ with a learning rate of $10^{-5}$ and one seed. TSC is ahead by $+0.69$ points
at step 250 and by $+0.49$ on average over twelve checkpoints. The same configuration and
rate give $+1.59$ on 1.7B. The 4B model already solves the prompts that uniform sampling
serves 54\% of the time, against 39\% for 1.7B, so it has less to gain from a different
choice of prompts (see details in Appendix~\ref{app:4b}).

\subsection{Code Generation with Unit Tests}\label{sec:code}

We repeated the comparison on Python programming problems from KodCode-Light-RL-10K
\citep{xu2025kodcode}, with the same 1.7B model, training recipe and TSC constants. The
reward is 1 if a program passes all unit tests of its problem and 0 otherwise. We choose the
learning rate of each method on 483 held-out KodCode problems and use HumanEval+ and MBPP+
(version 0.2.0, 378 problems) \citep{liu2023evalplus} only as test sets. This pool has more prompts that give zero-variance
groups at eight rollouts than the math pool, 68\% against 51\%. We trained each method with
two seeds in two configurations, $\shape{8}{8}$ for 500 steps and $\shape{32}{4}$ for 250
steps, so that both configurations use 32{,}000 rollouts. Each run keeps a checkpoint every 50
steps. (See Appendix~\ref{app:code} for details.)

\begin{figure}[t]
\centering
\begin{tikzpicture}
\begin{axis}[
  width=.72\linewidth, height=5.6cm,
  xlabel={training rollouts (64 per step, 500 steps)},
  ylabel={held-out avg@16 (\%)},
  xmin=-800, xmax=33000, ymin=65.6, ymax=72.8, ytick={66,67,68,69,70,71,72},
  xtick={0,6400,12800,19200,25600,32000}, xticklabels={0,{6{,}400},{12{,}800},{19{,}200},{25{,}600},{32{,}000}},
  scaled x ticks=false,
  legend pos=north west, legend cell align=left, legend style={font=\scriptsize},
  tick label style={font=\small}, label style={font=\small},
  grid=major, grid style={line width=.1pt, draw=gray!18},
]
\addplot[gray, dashed, thin, forget plot] coordinates {(-800,70.945) (33000,70.945)};
\addplot[only marks, mark=o, mark size=1.5pt, cUnif, forget plot] coordinates {(3200,66.64) (6400,68.31) (9600,68.27) (12800,69.46) (16000,69.86) (19200,70.03) (22400,70.26) (25600,69.93) (28800,69.98) (32000,69.68)};
\addplot[only marks, mark=o, mark size=1.5pt, cUnif, forget plot] coordinates {(3200,68.45) (6400,68.12) (9600,67.82) (12800,68.67) (16000,69.86) (19200,69.91) (22400,71.29) (25600,71.96) (28800,71.66) (32000,71.51)};
\addplot[only marks, mark=square, mark size=1.4pt, cOurs, forget plot] coordinates {(3200,67.60) (6400,67.53) (9600,70.11) (12800,70.51) (16000,71.04) (19200,70.92) (22400,71.11) (25600,71.33) (28800,70.99) (32000,71.66)};
\addplot[only marks, mark=square, mark size=1.4pt, cOurs, forget plot] coordinates {(3200,68.19) (6400,67.51) (9600,68.53) (12800,69.20) (16000,70.98) (19200,71.27) (22400,72.09) (25600,72.26) (28800,72.42) (32000,72.06)};
\addplot[cUnif, thick, mark=*, mark size=1.7pt] coordinates {(0,66.08) (3200,67.545) (6400,68.215) (9600,68.045) (12800,69.065) (16000,69.86) (19200,69.97) (22400,70.775) (25600,70.945) (28800,70.82) (32000,70.595)};
\addplot[cOurs, thick, mark=square*, mark size=1.7pt] coordinates {(0,66.08) (3200,67.895) (6400,67.52) (9600,69.32) (12800,69.855) (16000,71.01) (19200,71.095) (22400,71.60) (25600,71.795) (28800,71.705) (32000,71.86)};
\draw[cOurs!80!black, thin, dotted] (axis cs:16000,65.6) -- (axis cs:16000,71.01);
\node[font=\scriptsize, text=cOurs!80!black, anchor=north west, align=left] at (axis cs:16400,67.2) {TSC reaches 70.9\\with 16{,}000 rollouts};
\node[font=\scriptsize, text=gray!70!black, anchor=north west] at (axis cs:6800,70.92) {best of uniform: 70.9};
\legend{{uniform sampling, mean of 2 seeds}, {TSC, mean of 2 seeds}}
\end{axis}
\end{tikzpicture}
\caption{\textbf{Held-out accuracy on the code task against the number of training rollouts}
(Qwen3-1.7B, $\shape{8}{8}$, learning rate $10^{-5}$, a checkpoint every 50 steps). Lines are
the means of two seeds per method, and open markers are the individual seeds. The dashed line
marks the highest mean accuracy of uniform sampling, 70.9\% after 25{,}600 rollouts.}
\label{fig:code}
\end{figure}
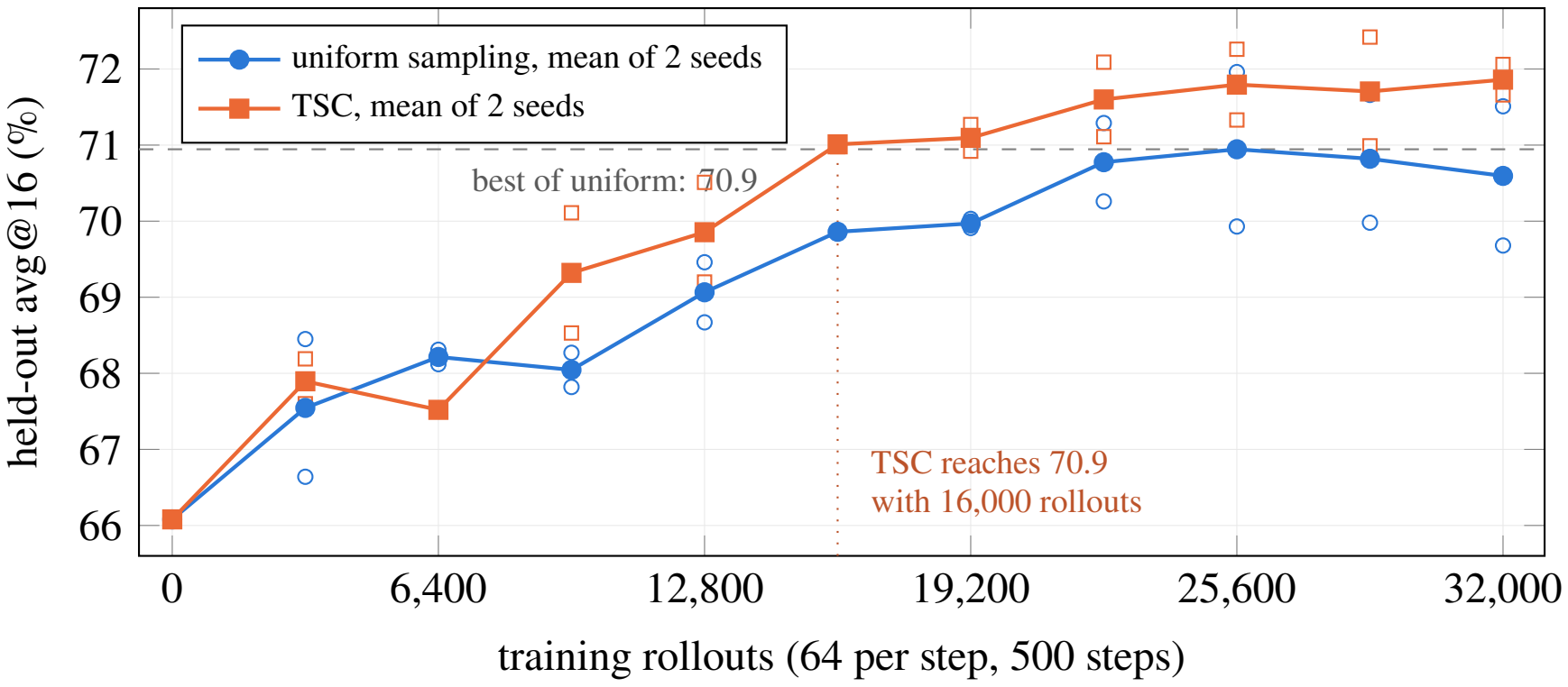

\paragraph{TSC is ahead from early in training.} Figure~\ref{fig:code} shows the held-out
accuracy at $\shape{8}{8}$, averaged over two seeds. TSC is ahead at every checkpoint from
9{,}600 rollouts (step 150) on, by 0.8 to 1.3 points. Before that, TSC is ahead at 3{,}200
rollouts and behind at 6{,}400.

\paragraph{TSC increases the gain from training by about 28\%.} After 32{,}000 rollouts
(step 500), TSC reaches 71.9\% and uniform sampling reaches 70.6\%, a difference of 1.3
points. Both TSC seeds score above both uniform seeds. The untrained model scores 66.1\%, so
training adds 4.5 points with uniform sampling and 5.8 points with TSC, about 28\% more.
Training moves accuracy by only a few points on this pool with either method. This is in
line with the average gains of 1.7 to 2.7 points across coding benchmarks reported for GRPO
on KodCode data with 7B models \citep{xu2025kodcode}. At $\shape{32}{4}$, after the same
32{,}000 rollouts (step 250), TSC is ahead by 0.8 points (70.5\% against 69.8\%), and again
both TSC seeds score above both uniform seeds.

\paragraph{TSC reaches the best accuracy of uniform sampling with 63\% of the rollouts.}
The highest accuracy of uniform sampling is 70.9\%, after 25{,}600 rollouts. TSC first
reaches this accuracy after 16{,}000 rollouts, which is 63\% of that budget.

\paragraph{About twice as many of TSC's rollouts produce a gradient.} TSC lowers the share
of zero-variance groups from 63\% to 23\% at $\shape{8}{8}$ and from 72\% to 36\% at
$\shape{32}{4}$ (Table~\ref{tab:code} in Appendix~\ref{app:code}). Both methods generate
32{,}000 rollouts in each configuration. Under uniform sampling, 37\% and 28\% of these
rollouts are in groups that produce a gradient. Under TSC, the shares are 77\% and 64\%,
which is 2.1 and 2.3 times as many.

\paragraph{On the additional test sets, TSC is ahead on HumanEval+ and behind on MBPP+.} We chose the
learning rate on the held-out set, so the two additional test sets give an independent check.
Table~\ref{tab:code} compares the methods after 32{,}000 rollouts. In both configurations,
TSC is ahead on HumanEval+, by 1.0 and 0.4 points, and behind on MBPP+, by 1.5 and 0.6
points. These differences are within the spread between seeds. For example, the two TSC
seeds at $\shape{8}{8}$ differ by 3.8 points on HumanEval+. The two additional test sets also differ in
format. Every held-out KodCode prompt and every HumanEval+ prompt gives the signature of the
function to write. An MBPP+ prompt gives a one-sentence task and one example test instead.
TSC may fit the format of its training problems more closely, which would help on HumanEval+
and not on MBPP+.

\subsection{A Research Agent with Rubric Rewards}\label{sec:agent}

The third task has a graded reward and long outputs. It is the task of
FrontierFinance~\citep{zhang2026frontierfinance}, a public benchmark of 220 expert-written
financial research questions that are graded against 11{,}543 source-attributed rubric
items. The policy is Nemotron 3.5 Lightning, a model with 30B parameters of which 3B are
active, and it acts as the research agent. It searches financial
filings, transcripts, market data and the web over many tool calls and then writes a report.
Each question has its own rubric, and a language-model judge scores the report by the share
of the rubric's criteria that it meets. We train it with the recipe of Samaya Sterling
\citep{koyluoglu2026sterling} and experiment with its last stage, online RL with GRPO, on 352
training questions (see Appendix~\ref{app:agent} for details of the recipe). We evaluate on 100 held-out
questions with a panel of three language-model judges.

\paragraph{Zero-variance groups are rare on this task.} Among 352 groups of the agent's own
rollouts, only 1.1\% have zero variance, against 51\% on math and 68\% on code. Two reports
almost never receive exactly the same partial credit. The reward spread is also similar
across the pool. When we divide the questions into five bins of equal size by mean reward,
the hardest bin has a mean spread of 0.07 and the other four have 0.11 to 0.13. The signal
that TSC selects on is therefore weaker here than on math.

\paragraph{Online training.} We then ran TSC online with fresh rollouts, using the recipe of
the uniform-sampling baseline. Each training step has 128 rollouts. Uniform sampling serves
each of the 352 training questions once, in random order, which takes 22 steps. We adapt TSC to
the graded reward in two ways. The bins are quantiles of each question's current mean
reward, so a question moves between bins as the policy improves. The Thompson draw is
multiplied by a weight that falls with the distance of a bin from the middle bin, so most
questions come from medium difficulty. Each question is served at most twice.

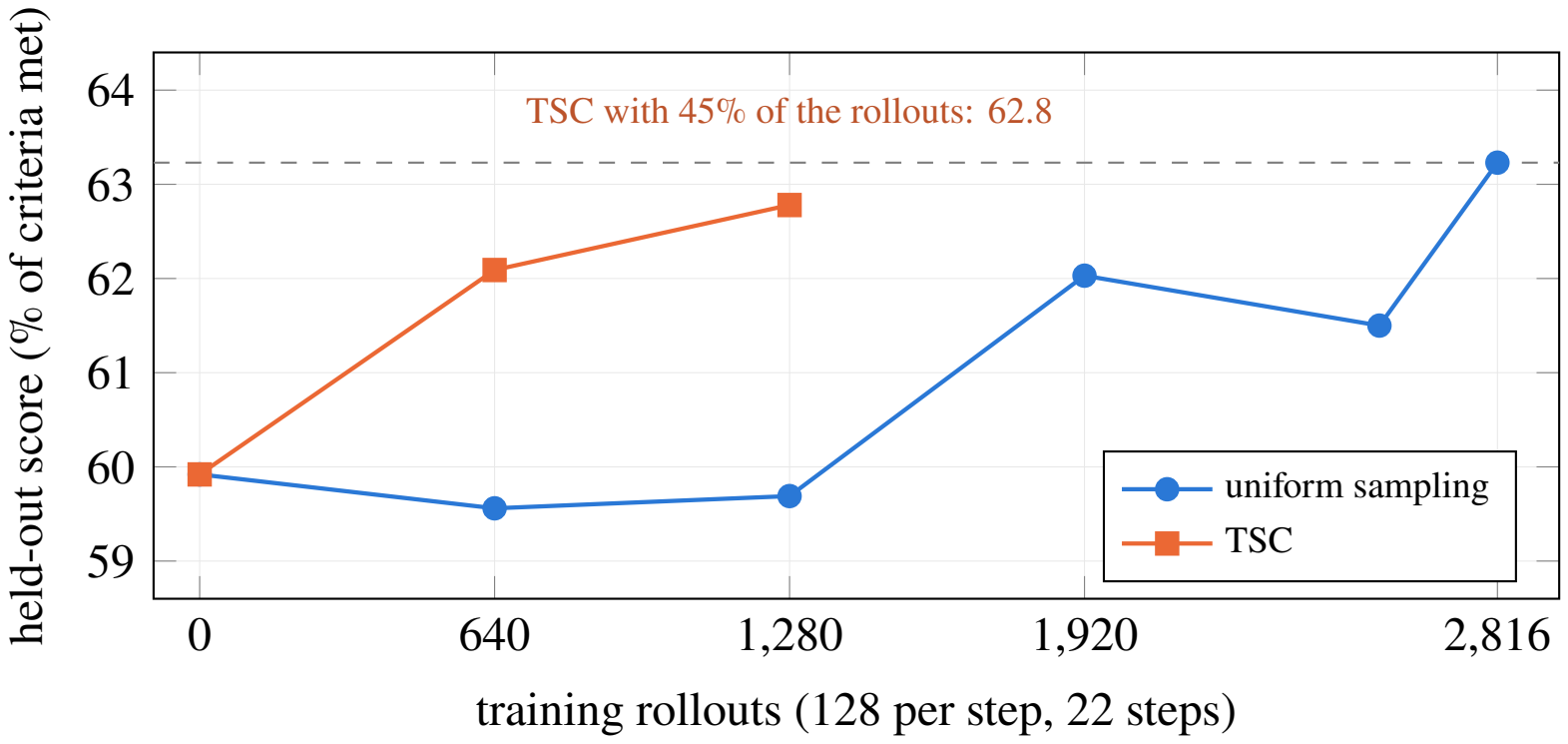
\begin{figure}[t]
\centering
\begin{tikzpicture}
\begin{axis}[
  width=.66\linewidth, height=5.2cm,
  xlabel={training rollouts (128 per step, 22 steps)},
  ylabel={held-out score (\% of criteria met)},
  xmin=-100, xmax=2950, ymin=58.6, ymax=64.4, ytick={59,60,61,62,63,64},
  xtick={0,640,1280,1920,2816}, xticklabels={0,640,{1{,}280},{1{,}920},{2{,}816}},
  scaled x ticks=false,
  legend pos=south east, legend cell align=left, legend style={font=\scriptsize},
  tick label style={font=\small}, label style={font=\small},
  grid=major, grid style={line width=.1pt, draw=gray!18},
]
\addplot[gray, dashed, thin, forget plot] coordinates {(-100,63.23) (2950,63.23)};
\addplot[cUnif, thick, mark=*, mark size=1.9pt] coordinates {(0,59.92) (640,59.56) (1280,59.69) (1920,62.03) (2560,61.50) (2816,63.23)};
\addplot[cOurs, thick, mark=square*, mark size=1.9pt] coordinates {(0,59.92) (640,62.09) (1280,62.78)};
\node[font=\scriptsize, text=cOurs!80!black, anchor=south, align=center] at (axis cs:1280,63.45) {TSC with 45\% of the rollouts: 62.8};
\legend{uniform sampling, TSC}
\end{axis}
\end{tikzpicture}
\caption{\textbf{Held-out score of the research agent against the number of training
rollouts} (one run per method, checkpoints every five steps). The dashed line marks the score of
uniform sampling after all 2{,}816 rollouts. TSC is shown up to the checkpoint selected by
early stopping, after 1{,}280 rollouts.}
\label{fig:agent}
\end{figure}

\paragraph{TSC reaches the final score of uniform sampling with 45\% of the rollouts.} The
score of uniform sampling stays near its starting value for the first 1{,}280 rollouts (step 10)
and then rises to 63.2\% after all 2{,}816 (step 22, Figure~\ref{fig:agent}). TSC rises
earlier. It scores 62.1\% after 640 rollouts, a score that uniform sampling reaches only at the
checkpoint after 1{,}920 rollouts (62.0\%). After 1{,}280 rollouts, TSC scores 62.8\%. The
paired difference between this checkpoint and the final checkpoint of uniform sampling is
$-0.45$ points \ci{-3.69}{+2.55} (paired bootstrap over the held-out questions), so TSC gives a
model with a similar score from 45\% of the rollouts. The leads of TSC over uniform sampling at steps 5 and
10 are 2 to 3 points, and their paired intervals include zero.

\paragraph{The saving requires early stopping.} Training TSC beyond step 10 did not improve
its score. On this task, the benefit of TSC is the saving in rollouts, and it is obtained by
stopping at the best checkpoint on a held-out set.

\section{Discussion}\label{sec:task}

\paragraph{When to expect a gain.} The three tasks suggest two measurements to make before
using a curriculum of this kind. The first is the share of zero-variance groups under
uniform sampling. It is about one half on math and about two thirds on code, where TSC raised
the accuracy and saved rollouts. It is 1\% on the research agent (with graded non-binary rewards), where TSC saved rollouts but
did not raise the final score. The second is how much the reward spread differs between
prompts, because any selector based on variance relies on this difference. With a binary reward, the spread of a group depends only on the solve rate of its prompt and differs widely
across a pool. With partial credit, this does not need to hold. If even a poor answer earns some
credit, zero-variance groups become rare, but the spread also becomes similar across prompts
(Appendix~\ref{app:betabin}). This is what we measured on the agent task. A graded reward can
keep this difference if wrong answers earn little credit. For example, a code reward could be
the fraction of unit tests passed, provided that a wrong program usually passes few of them. Within a task, the batch configuration matters as well. TSC helped when the prompts of a step came from a varied mix of bins,
measured as an entropy above about 1.14 bits (Section~\ref{sec:diversity}). This held in all six
configurations with at least eight prompts per step and in one of the four configurations with
four prompts.

\paragraph{The cost of the bins.} TSC needs a solve count for every prompt. One pass of the
untrained model over the representative math pool that we use here, with eight rollouts per prompt,
generates 303{,}760 rollouts. This cost is paid once and
shared by every run on the pool. DeepScaleR provides its own difficulty labels, measured with
a stronger reference model, but they may not be a good substitute. They correlate $-0.64$ with
the solve rate of the 1.7B model and identify only about a third of the prompts that give
zero-variance groups for this model. Labels that track the model's own solve rate more
closely may still replace the pass of the untrained model. Examples are a cheaper estimate
from a few rollouts per prompt, a difficulty predictor trained on the prompt text, or labels
that a task provides with its data (Section~\ref{sec:future}).

\section{Limitations and Future Work}\label{sec:limits}

\subsection{Limitations}

\paragraph{Seeds and noise.} We use two seeds for most settings. The research agent experiment and
the 4B experiments use one run per method. Some pairs of seeds differ by much more than the
0.80-point standard deviation of a single run (Appendix~\ref{app:repro}). More than two seeds
per setting would lower the noise of each comparison.

\paragraph{Models and training.} On the math task, all ten configurations were run only on the
1.7B model. The 0.6B experiments cover three configurations, and the 4B comparison uses one
configuration. All
three models belong to one family. All training uses LoRA with one recipe per task. When we
double the LoRA rank at $\shape{8}{8}$, uniform sampling is unchanged and the gain of TSC
falls from $+2.01$ to $+0.48$ (Appendix~\ref{app:rank}), so the size of the gain depends on
the adapter. We did not test full fine-tuning.

\paragraph{Other methods and tasks.} The comparison with other selection methods uses one
configuration and one model. We did not run the per-prompt selectors GRESO, KGPS and MoPPS,
for the reason given in Section~\ref{sec:related}, or VCRL and online difficulty filtering,
which filter after generation as DAPO does. Besides math, we tested two tasks, and all
experiments use GRPO. More tasks and other learning algorithms would show how TSC affects
other datasets and algorithms beyond GRPO.

\subsection{Future Work}\label{sec:future}

\paragraph{Updating the bins during training.} In the math and code experiments, a prompt
keeps the bin that the untrained model assigned to it. The beliefs follow the change in the
reward spread of each bin. A prompt that the model has learned to solve, however, stays in its
original bin, so each bin becomes more mixed as training proceeds. Over 1{,}000 steps, the share of zero-variance groups under TSC rises from 10.7\% to 14.2\%
(Appendix~\ref{app:durability}), which may reflect this mixing. A direct extension measures the
solve counts again after a fixed number of steps, using the latest checkpoint, and moves each prompt to its new bin. The
rollouts generated during training provide new solve counts at no cost, but only for the
prompts that were served, and the average prompt is served less than once in a run. A second
pass over the full pool costs 303{,}760 rollouts, so a practical version would measure again
only a subset of the pool or only the bins that TSC currently serves. The research agent experiment
already updates the bins during training, because it assigns each question to a bin by its
current mean reward. With 352 training questions, each bin holds about 70 questions, and we
wanted each question to be in the bin that matches the current policy. The rollouts of each
step provide the new estimates at no extra cost. We did not compare these updated bins with fixed bins. The effect of updating the bins should be largest in runs much longer
than 250 steps.

\paragraph{Bins over topic and difficulty.} The nine bins use difficulty alone. We ran TSC on
21 clusters of topic and difficulty, and it scored lower than on the nine bins (41.01\%
against 42.25\%, Table~\ref{tab:comparators}). These clusters have only three difficulty
levels, so the comparison does not show whether topic helps when all nine levels are kept.
Crossing the nine bins with the seven topics gives 63 bins. A 250-step run at $\shape{8}{8}$
has 2{,}000 groups, which is 32 observations per bin on average against 222 with nine bins.
The beliefs would therefore need to share information across bins, for example with one
belief per difficulty level and a correction per topic. Bins over topic would let TSC
detect a topic on which the model has stopped improving, and control the mix of topics in a
step. TSC currently takes 89\% of its prompts from three bins and leaves the topics within
them to chance. For pools without topic labels, clusters of prompt embeddings can serve as
topics, as the seven topic clusters provided with DeepScaleR do.

\paragraph{Cheaper bins.} The labels do not need to be precise. Measuring with
sixteen rollouts instead of eight moves 34\% of the prompts to a different bin and leaves the
mean spread and the ranking of the bins almost unchanged (Appendix~\ref{app:kalman}). Fewer
rollouts per prompt may be sufficient. Another option is to measure a part of the
pool and to predict the bins of the remaining prompts from the prompt text or from the
difficulty labels provided with the dataset. Cheaper bins
would also make it affordable to update them during training.

\paragraph{Prompts that the model never solves.} TSC serves almost no prompts from the
never-solved bin, which holds 34.3\% of the pool, and neither method solves many problems
that the untrained model could not solve (Section~\ref{sec:coverage}). Larger groups alone
are unlikely to make these prompts useful, because each doubling of the rollouts per prompt
removes only about a fifth of the zero-variance groups. Further experiments can test a group size
that differs by bin, and easier versions of these prompts, for example with a hint or a
partial solution.

\paragraph{A selection signal for graded rewards.} On the agent task, the reward spread is
similar across bins, so it separates the questions only weakly. Each criterion of a rubric is
either met or not met in a rollout. One candidate is the spread of each criterion across the
rollouts of a question, averaged over the criteria of the rubric. Rollouts that meet
different criteria can receive the same total score, so this signal may separate the
questions better than the spread of the total score. Another
candidate is the change in a bin's mean reward over recent steps, which measures whether the
model is still improving on that bin. Either signal can replace $\hat\sigma_p$ in the belief
update, with the observation noise $R$ measured again for the new signal.

\section{Conclusion}\label{sec:conclusion}

About half of the groups that GRPO generates on a standard math pool have zero variance, and
a curriculum over difficulty bins removes most of them before any rollout is generated. We
compared such a curriculum, TSC, with uniform sampling across ten batch configurations ($P$ prompts per
step and $G$ rollouts per prompt), with the learning rate tuned for each method. TSC reaches a higher best accuracy, and the gain
comes from problems that the model solves more reliably. The gain appears when a step has
enough prompts. Across these configurations and learning rates, TSC is the more
efficient method for reaching high accuracy. In particular, compared with DAPO, which drops
zero-variance groups after they are generated, TSC reaches the same accuracy and generates
about half as many rollouts. On two other tasks, TSC also saves rollouts. On the code task we
study, TSC increases the accuracy gain from training by about 28\% and reaches the best
held-out accuracy of uniform sampling with 63\% of the rollouts. On a rubric-graded research agent, where
zero-variance groups are rare and the reward spread is similar across bins, TSC reaches the
final score of uniform sampling with 45\% of the rollouts. The reward spread of
the bins is a practical signal for choosing prompts before any rollout is generated. We
show that a curriculum that follows this signal can save rollouts and increase accuracy and
reliability.

\appendix

\section{Further Results}\label{app:further}

\subsection{Results for All Configurations}\label{app:grid}

\begin{figure}[tbp]
\centering
\begin{tikzpicture}
\begin{axis}[
  width=\linewidth, height=6.0cm,
  xlabel={configuration $P{\times}G$ (ordered by rollouts per step, then by rollouts per prompt)},
  ylabel={avg@16},
  xmin=0.4, xmax=10.6, ymin=35.5, ymax=43.8,
  xtick={1,2,3,4,5,6,7,8,9,10},
  xticklabels={4$\times$4,8$\times$4,4$\times$8,16$\times$4,8$\times$8,4$\times$16,32$\times$4,16$\times$8,8$\times$16,4$\times$32},
  x tick label style={font=\scriptsize},
  ytick={36,37,38,39,40,41,42,43},
  tick label style={font=\small}, label style={font=\small},
  legend pos=south east, legend cell align=left, legend style={font=\scriptsize},
  grid=both, grid style={line width=.1pt, draw=gray!16},
]
\addplot[gray, dashed, thin, forget plot] coordinates {(0.4,37.54) (10.6,37.54)};
\addplot[only marks, mark=o, mark size=1.4pt, cUnif, opacity=0.5] coordinates {
(0.84,37.76) (0.84,38.22) (0.84,38.91) (0.84,39.06) (0.84,40.95) (0.84,36.89) (1.84,38.38) (1.84,39.39) (1.84,39.55) (1.84,39.84) (1.84,38.62) (2.84,38.44) (2.84,38.68) (2.84,40.25) (2.84,40.70) (2.84,38.42) (3.84,39.12) (3.84,39.31) (3.84,40.00) (3.84,40.48) (3.84,38.38) (4.84,38.88) (4.84,40.24) (4.84,39.98) (4.84,38.70) (4.84,39.55) (5.84,39.35) (5.84,40.80) (5.84,41.64) (5.84,40.12) (6.84,40.66) (6.84,40.87) (6.84,39.02) (7.84,40.93) (7.84,41.43) (7.84,39.89) (8.84,40.45) (8.84,41.45) (8.84,41.00) (8.84,39.99) (9.84,41.45) (9.84,41.59) (9.84,41.41)};
\addplot[only marks, mark=o, mark size=1.4pt, cOurs, opacity=0.5] coordinates {
(1.16,38.21) (1.16,38.93) (1.16,39.25) (1.16,39.59) (1.16,38.28) (2.16,39.41) (2.16,40.06) (2.16,40.77) (2.16,40.41) (3.16,39.86) (3.16,39.95) (3.16,40.49) (3.16,40.02) (4.16,39.84) (4.16,40.70) (4.16,41.12) (4.16,40.77) (5.16,40.70) (5.16,40.29) (5.16,42.25) (5.16,42.08) (6.16,41.02) (6.16,41.28) (6.16,41.31) (6.16,40.30) (7.16,40.93) (7.16,42.94) (7.16,40.92) (7.16,38.09) (8.16,41.04) (8.16,42.30) (8.16,39.75) (8.16,35.94) (9.16,41.33) (9.16,41.32) (9.16,42.25) (9.16,39.54) (10.16,41.72) (10.16,42.53) (10.16,42.07) (10.16,41.67)};
\addplot[cUnif, thick, mark=*, mark size=2.2pt] coordinates {
(1,40.95) (2,39.84) (3,40.70) (4,40.48) (5,40.24) (6,41.64) (7,40.87) (8,41.43) (9,41.45) (10,41.59)};
\addplot[cOurs, thick, mark=square*, mark size=2.2pt] coordinates {
(1,39.59) (2,40.77) (3,40.49) (4,41.12) (5,42.25) (6,41.31) (7,42.94) (8,42.30) (9,42.25) (10,42.53)};
\legend{uniform (all rates), TSC (all rates), uniform (best rate), TSC (best rate)}
\end{axis}
\end{tikzpicture}
\caption{\textbf{All configurations and learning rates on Qwen3-1.7B.} One faint point per batch
configuration, sampler and learning rate (mean of two seeds), with each method's best rate at
each configuration connected. The dashed line is the untrained model (37.54).}
\label{fig:allscores}
\end{figure}
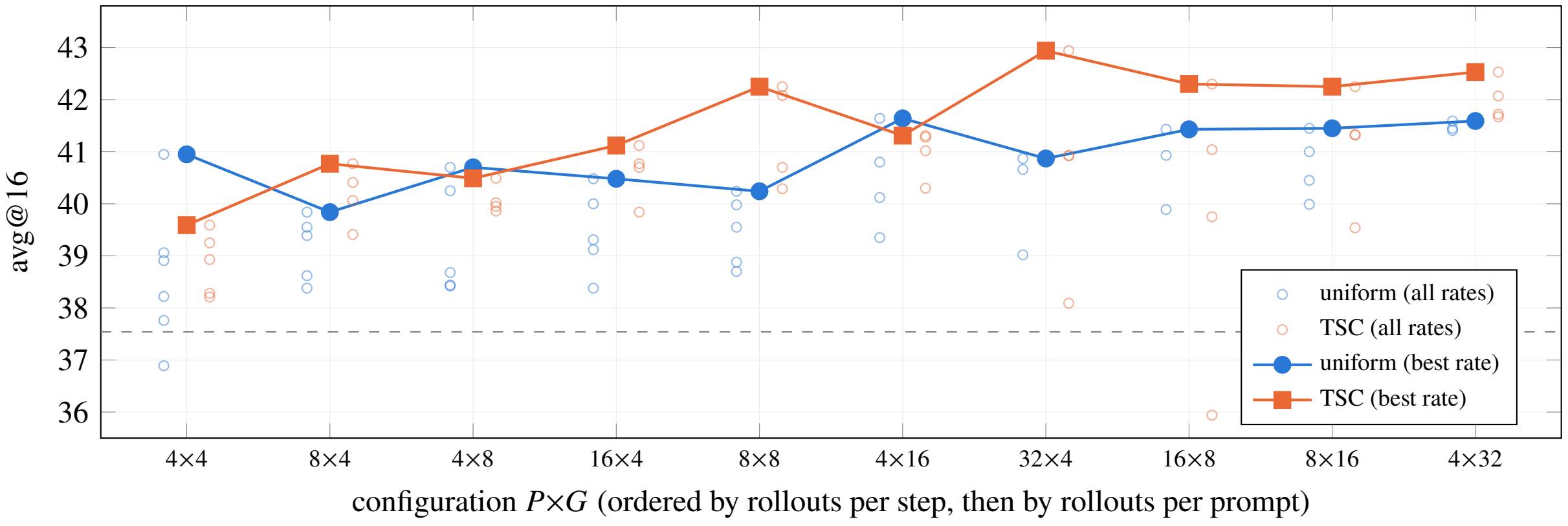

Figure~\ref{fig:master} shows all 168 runs of the grid, one marker for each combination of
configuration, learning rate and seed, against the number of rollouts the run generated.
Figure~\ref{fig:allscores} shows the accuracy of both methods for every configuration and
learning rate. Table~\ref{tab:grid} lists the results at the best rates.

\begin{figure}[tbp]
\centering
\begin{tikzpicture}
\begin{axis}[
  width=0.92\linewidth, height=7.0cm,
  xmode=log, log basis x=2,
  xmin=2.9, xmax=44, ymin=32.5, ymax=44.8,
  xtick={4,8,16,32}, xticklabels={4k,8k,16k,32k},
  xlabel={rollouts generated over 250 steps ($P\cdot G\cdot 250$)},
  ylabel={avg@16 (held-out DeepScaleR)},
  ytick={33,35,37,39,41,43},
  tick label style={font=\small}, label style={font=\small},
  legend cell align=left, legend columns=4, legend style={font=\scriptsize, at={(0.5,-0.2)}, anchor=north, draw=gray!40, /tikz/every even column/.append style={column sep=6pt}},
  grid=major, grid style={line width=.1pt, draw=gray!16},
  clip=true,
]
\addplot[fill=figConBg, draw=none, forget plot] coordinates {(2.9,40.95) (16,40.95) (16,41.64) (44,41.64) (44,44.8) (2.9,44.8)} -- cycle;
\node[font=\scriptsize\sffamily, text=figCon!70!black, anchor=north west, align=left] at (axis cs:3.05,44.6)
  {above every uniform setting\\ at equal or lower cost};
\addplot[gray, dashed, thin, forget plot] coordinates {(2.9,37.54) (44,37.54)};
\node[font=\scriptsize, text=gray, anchor=south] at (axis cs:11.3,37.54) {untrained model};
\addplot[only marks, mark=o, mark size=1.5pt, cUnif, opacity=0.55] coordinates {(13.977,39.65) (13.977,40.35) (13.977,39.04) (13.977,39.21) (13.977,40.31) (13.977,40.66) (13.977,39.04) (13.977,37.71) (13.977,39.65) (13.977,38.98) (29.548,41.70) (29.548,40.16) (29.548,41.84) (29.548,41.02) (29.548,40.33) (29.548,39.46) (26.446,40.03) (26.446,41.30) (26.446,40.80) (26.446,40.94) (26.446,38.33) (26.446,39.71) (17.448,41.58) (17.448,41.71) (17.448,39.35) (17.448,39.35) (17.448,39.51) (17.448,40.74) (17.448,40.99) (17.448,40.61) (36.886,42.10) (36.886,40.81) (36.886,41.98) (36.886,41.21) (36.886,40.91) (36.886,41.92) (3.904,38.39) (3.904,38.04) (3.904,41.65) (3.904,40.25) (3.904,38.91) (3.904,38.90) (3.904,36.07) (3.904,37.70) (3.904,39.04) (3.904,39.08) (3.904,37.86) (3.904,37.66) (8.253,38.11) (8.253,38.77) (8.253,41.32) (8.253,40.09) (8.253,38.36) (8.253,39.00) (8.253,38.04) (8.253,38.80) (8.253,40.21) (8.253,40.29) (33.014,42.00) (33.014,40.89) (33.014,40.05) (33.014,41.95) (33.014,40.23) (33.014,39.76) (33.014,40.76) (33.014,40.14) (7.387,38.84) (7.387,37.91) (7.387,40.04) (7.387,39.64) (7.387,39.92) (7.387,38.86) (7.387,39.02) (7.387,38.23) (7.387,39.21) (7.387,39.89) (15.617,38.92) (15.617,38.83) (15.617,39.64) (15.617,37.77) (15.617,40.29) (15.617,40.19) (15.617,38.85) (15.617,40.25) (15.617,40.10) (15.617,39.85)};
\addplot[only marks, mark=square, mark size=1.4pt, cOurs, opacity=0.6] coordinates {(14.672,40.55) (14.672,40.86) (14.672,41.21) (14.672,41.04) (14.672,40.65) (14.672,40.88) (14.672,39.46) (14.672,40.21) (31.017,40.70) (31.017,41.38) (31.017,42.48) (31.017,42.11) (31.017,40.56) (31.017,38.94) (31.017,32.90) (31.017,38.98) (27.761,40.45) (27.761,41.41) (27.761,42.62) (27.761,43.25) (27.761,41.30) (27.761,40.55) (27.761,37.08) (27.761,39.11) (18.316,40.83) (18.316,41.74) (18.316,41.02) (18.316,41.60) (18.316,39.01) (18.316,41.58) (18.316,41.42) (18.316,40.62) (38.720,42.31) (38.720,42.75) (38.720,42.15) (38.720,41.99) (38.720,39.27) (38.720,44.08) (38.720,41.71) (38.720,41.73) (4.098,38.26) (4.098,38.16) (4.098,40.02) (4.098,39.15) (4.098,39.01) (4.098,38.85) (4.098,38.80) (4.098,37.76) (4.098,40.01) (4.098,38.49) (8.664,40.55) (8.664,40.44) (8.664,40.25) (8.664,39.48) (8.664,39.15) (8.664,40.88) (8.664,39.76) (8.664,40.15) (34.655,40.70) (34.655,41.96) (34.655,41.49) (34.655,41.15) (34.655,41.86) (34.655,42.64) (34.655,38.38) (34.655,40.70) (7.754,40.69) (7.754,40.85) (7.754,39.85) (7.754,38.96) (7.754,40.55) (7.754,40.27) (7.754,41.35) (7.754,38.77) (16.393,40.25) (16.393,40.33) (16.393,41.50) (16.393,43.01) (16.393,41.66) (16.393,42.49) (16.393,40.66) (16.393,40.75)};
\addplot[cUnif, very thick, const plot mark left, mark=*, mark size=2.3pt] coordinates {(4,40.95) (8,40.95) (16,41.64) (32,41.64)};
\addplot[cOurs, very thick, const plot mark left, mark=square*, mark size=2.3pt] coordinates {(4,39.59) (8,40.77) (16,42.25) (32,42.94)};
\addplot[only marks, mark=diamond*, mark size=2.8pt, cDapo] coordinates {(34.28,42.51)};
\addplot[only marks, mark=triangle*, mark size=2.8pt, figCon!80!black] coordinates {(15.2,41.57)};
\addplot[only marks, mark=triangle*, mark options={rotate=180}, mark size=2.8pt, figCon!80!black] coordinates {(16.8,41.22)};
\node[font=\scriptsize, text=cDapo, anchor=west] at (axis cs:35.5,42.51) {DAPO};
\legend{{uniform, single run (86)}, {TSC, single run (82)}, {uniform, efficient frontier}, {TSC, efficient frontier}, {DAPO at $8{\times}8$}, {static headroom at $8{\times}8$}, {offline filtering at $8{\times}8$}}
\end{axis}
\end{tikzpicture}
\caption{\textbf{Every Qwen3-1.7B training run, by generation cost and accuracy.} Each small
marker is one run, a combination of batch configuration $P{\times}G$, learning rate and seed
(168 runs). Runs with the same number of rollouts are spread horizontally by configuration. The
thick lines are the efficient frontiers: the best two-seed mean each method reaches at a given
number of rollouts or fewer, with its learning rate tuned for each configuration. The shaded
region contains results better than every uniform setting at equal or lower cost. The TSC
frontier enters it at 16k rollouts (42.25) and 32k rollouts (42.94). A single run has a standard deviation of about
0.8 points (Appendix~\ref{app:repro}), so some single runs of both methods fall in the region. The diamond
is DAPO, the upward triangle is static headroom and the downward triangle is offline
filtering. These are the other selection methods of Section~\ref{sec:comparators} at
$8{\times}8$ (two-seed means at their best rates). Each is placed at the number of rollouts it
generates. Static headroom and offline filtering both generate 16k rollouts and are drawn side
by side.}
\label{fig:master}
\end{figure}
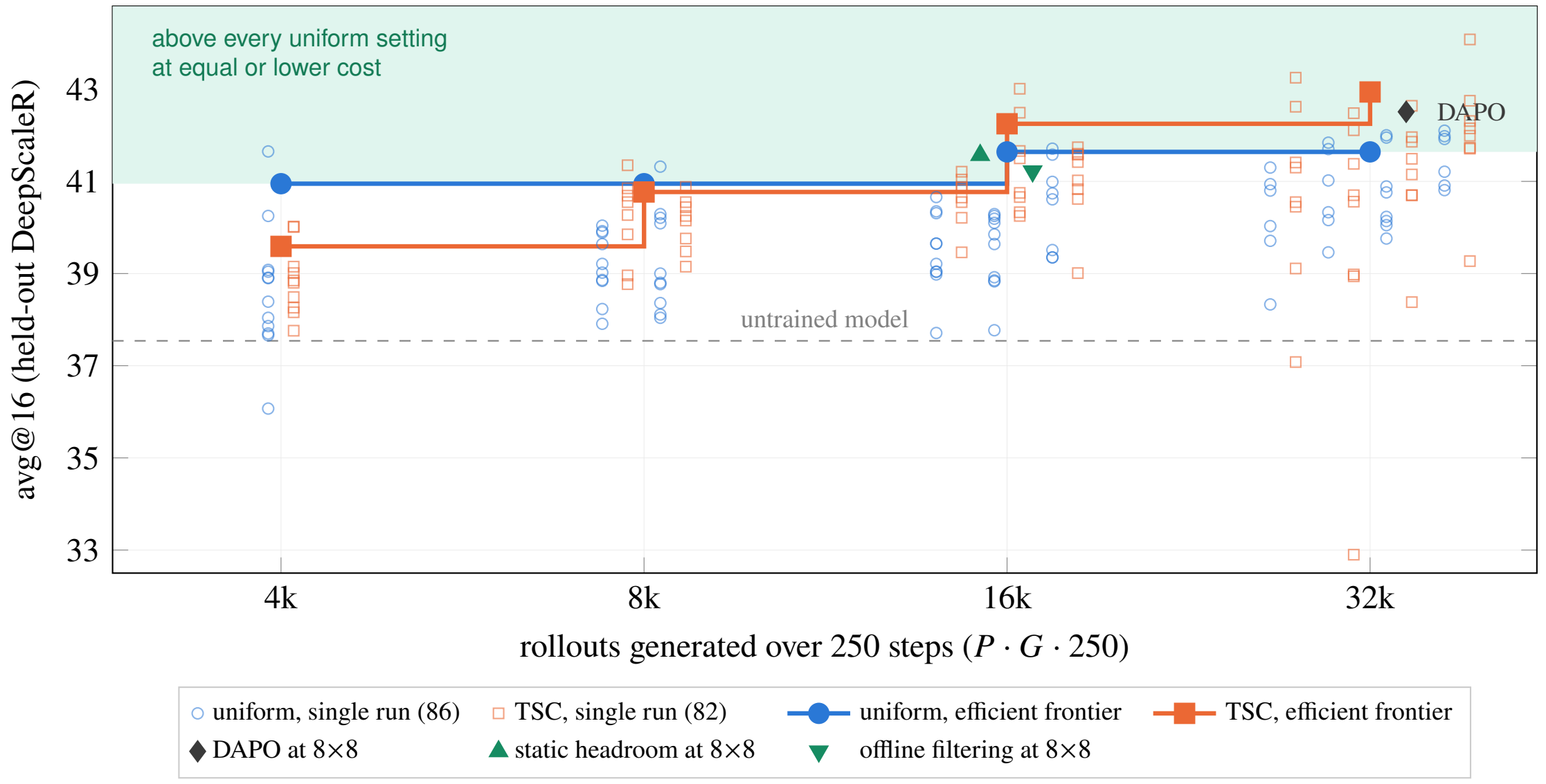

\subsection{Accuracy by Problem Difficulty}\label{sec:terciles}

We divide the 500 held-out problems into three equal groups by how often the untrained 1.7B
model solves them. The untrained avg@16 is 84.9\% on the easy group, 25.9\% on the medium
group and 1.2\% on the hard group. Table~\ref{tab:terciles} gives the gain of TSC on each
group. Medium problems gain the most, by $+1.02$ points on average over all configurations
and by $+2.24$ over the six configurations with at least eight prompts per step.

\begin{table}[tbp]
\centering\small
\begin{tabular}{lrrrr}
\toprule
 & easy & medium & hard & all\\
\midrule
all ten configurations (mean) & $+0.28$ & $+1.02$ & $+0.61$ & $+0.64$\\
configurations with $P\ge8$ (mean of six) & $+0.32$ & $+2.24$ & $+1.12$ & $+1.22$\\
\bottomrule
\end{tabular}
\caption{\textbf{Gain of TSC by difficulty of the held-out problems} (TSC minus uniform
sampling, avg@16, 250 steps). The three groups are equal thirds of the held-out set ordered
by the solve rate of the untrained 1.7B model.}
\label{tab:terciles}
\end{table}

\subsection{Sampler Allocation During Training}\label{sec:dynamics}

\begin{figure}[tbp]
\centering
\begin{minipage}[t]{.49\linewidth}\centering
\begin{tikzpicture}
\begin{axis}[
  width=\linewidth, height=4.8cm, ybar, bar width=3pt,
  xlabel={bin (solve count of the untrained model, out of 8)}, ylabel={share of draws (\%)},
  ymin=0, ymax=50, xtick={0,...,8}, enlarge x limits=0.07,
  legend pos=north east, legend cell align=left, legend style={font=\scriptsize},
  tick label style={font=\small}, label style={font=\small},
  grid=major, grid style={line width=.1pt, draw=gray!18},
]
\addplot[fill=cUnif, draw=cUnif] coordinates {(0,34.3) (1,12.3) (2,8.5) (3,6.7) (4,5.5) (5,5.3) (6,5.0) (7,5.3) (8,17.1)};
\addplot[fill=cOurs, draw=cOurs] coordinates {(0,0.2) (1,0.7) (2,6.7) (3,33.0) (4,42.1) (5,14.2) (6,2.2) (7,0.7) (8,0.1)};
\addplot[fill=white, draw=cOurs] coordinates {(0,0.2) (1,0.8) (2,7.4) (3,34.0) (4,39.2) (5,14.5) (6,3.6) (7,0.2) (8,0.1)};
\legend{uniform ($=$ pool), TSC $8{\times}8$, TSC $32{\times}4$}
\end{axis}
\end{tikzpicture}
\end{minipage}\hfill
\begin{minipage}[t]{.49\linewidth}\centering
\begin{tikzpicture}
\begin{axis}[
  width=\linewidth, height=4.8cm,
  xlabel={training step}, ylabel={share of draws (\%)},
  xmin=0, xmax=500, ymin=0, ymax=68, xtick={0,100,200,300,400,500},
  legend style={font=\scriptsize, at={(0.5,0.98)}, anchor=north, legend columns=3},
  tick label style={font=\small}, label style={font=\small},
  grid=both, grid style={line width=.1pt, draw=gray!18},
]
\addplot[red!75!black, thick, mark=triangle*] coordinates {(50,22.4) (150,35.1) (250,33.5) (350,38.9) (450,35.2)};
\addplot[cOurs, thick, mark=*] coordinates {(50,45.5) (150,39.6) (250,37.7) (350,38.4) (450,49.3)};
\addplot[figCon, thick, mark=square*] coordinates {(50,18.6) (150,15.2) (250,17.0) (350,12.1) (450,8.1)};
\legend{3 of 8 (harder), 4 of 8, 5 of 8 (easier)}
\end{axis}
\end{tikzpicture}
\end{minipage}
\caption{\textbf{Allocation of TSC over difficulty bins (1.7B, 500 steps, two seeds
pooled).} Left: share of draws by bin; uniform sampling matches the pool. Right: share of
each of the three middle bins in 100-step windows at the $8{\times}8$ configuration.}
\label{fig:dynamics}
\end{figure}
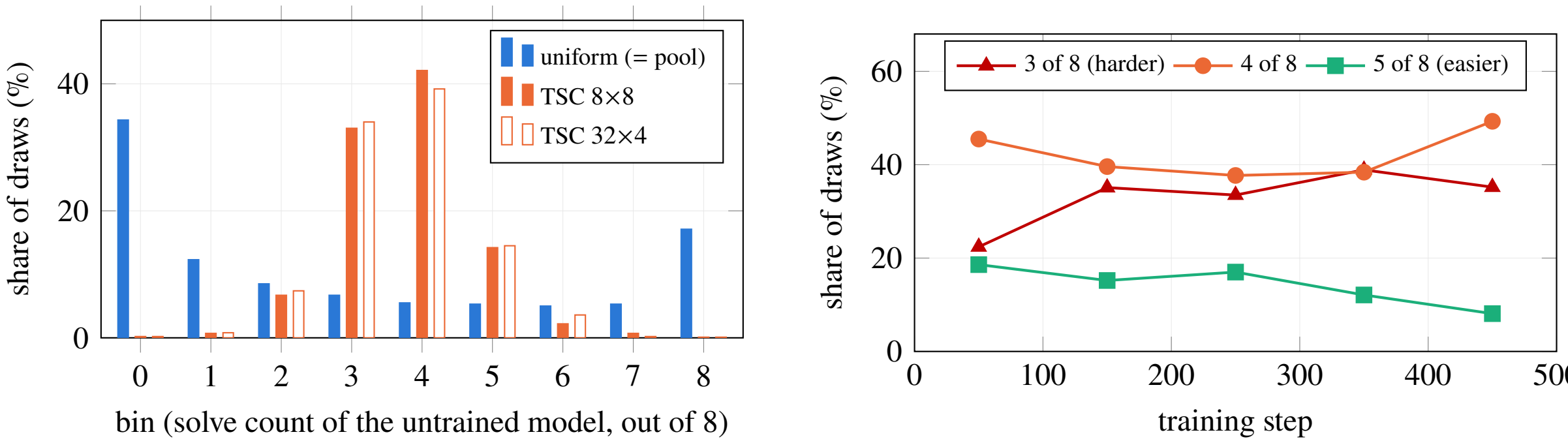

Over 500 steps at $\shape{8}{8}$, 89\% of TSC's draws come from the three bins in which the
untrained model solved three, four or five of eight rollouts (Figure~\ref{fig:dynamics}). These
bins hold 18\% of the pool. The never-solved and always-solved bins hold 51\% of the pool and
receive 0.3\% of the draws. Within the middle bins, the draws move toward harder prompts as
the model improves. The share of the five-of-eight bin falls from 19\% in the first 100 steps
to 8\% in the last 100, and the share of the three-of-eight bin rises from 22\% to 35\%. Over the same steps, the solve rate on the four-of-eight bin rises from 0.51 to 0.58. Because TSC
concentrates on a few bins, it serves the same prompts more often than uniform sampling. By step 500, the average prompt in the four-of-eight bin has been served 0.8 times at
$\shape{8}{8}$ and 3.0 times at $\shape{32}{4}$, against 0.1 and 0.4 times under uniform
sampling.

\subsection{Longer Training}\label{app:durability}\label{sec:durability}

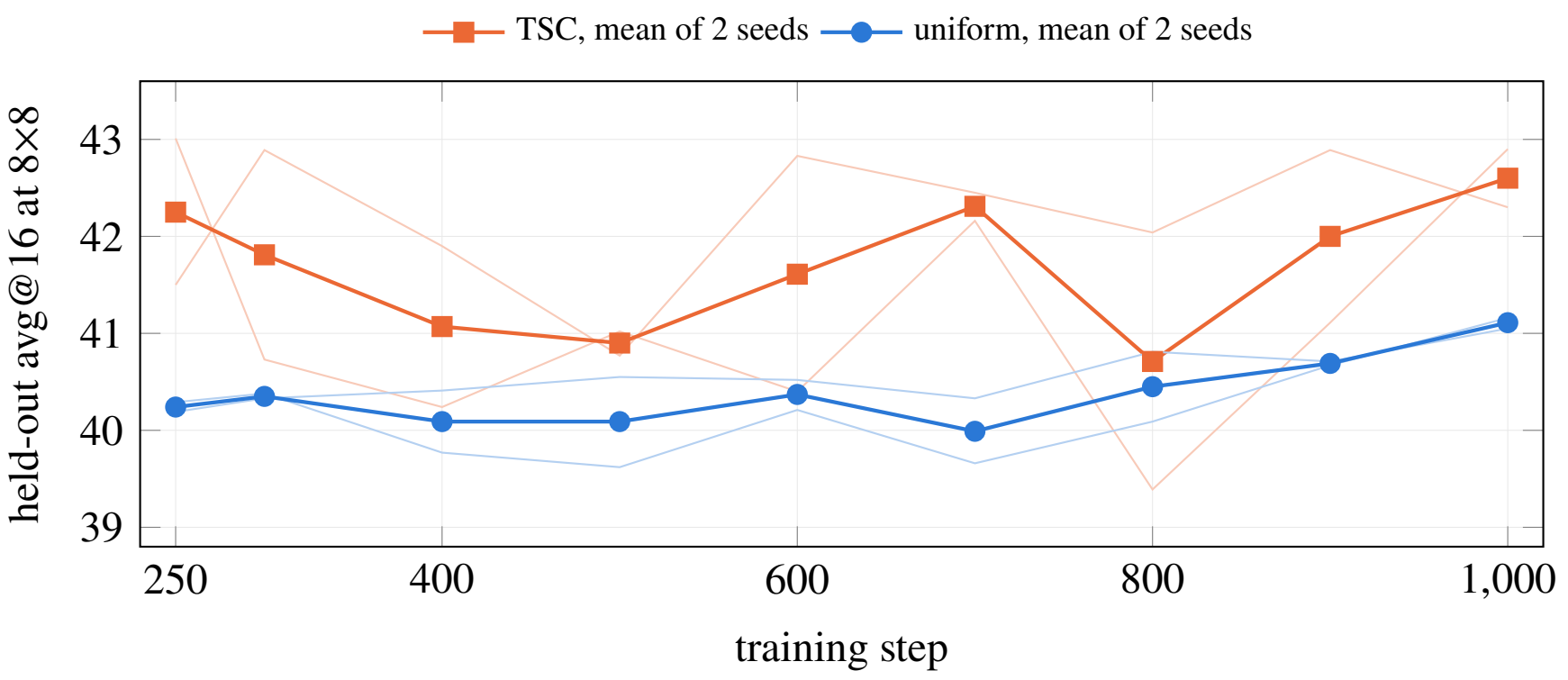
\begin{figure}[tbp]
\centering
\begin{tikzpicture}
\begin{axis}[
  width=.72\linewidth, height=5.0cm,
  xlabel={training step}, ylabel={held-out avg@16 at $8{\times}8$},
  xmin=230, xmax=1020, ymin=38.8, ymax=43.6,
  xtick={250,400,600,800,1000}, ytick={39,40,41,42,43},
  legend style={at={(0.5,1.03)}, anchor=south, font=\scriptsize, draw=none}, legend columns=3, legend cell align=left,
  tick label style={font=\small}, label style={font=\small},
  grid=major, grid style={line width=.1pt, draw=gray!18},
]
\addplot[cOurs!35, thin, mark=none, forget plot] coordinates {(250,41.50) (300,42.89) (400,41.90) (500,40.77) (600,42.83) (700,42.45) (800,42.04) (900,42.89) (1000,42.30)};
\addplot[cOurs!35, thin, mark=none, forget plot] coordinates {(250,43.01) (300,40.73) (400,40.24) (500,41.02) (600,40.40) (700,42.16) (800,39.39) (900,41.11) (1000,42.90)};
\addplot[cOurs, thick, mark=square*, mark size=1.8pt] coordinates {(250,42.25) (300,41.81) (400,41.07) (500,40.90) (600,41.61) (700,42.31) (800,40.71) (900,42.00) (1000,42.60)};
\addplot[cUnif!35, thin, mark=none, forget plot] coordinates {(250,40.29) (300,40.38) (400,39.77) (500,39.62) (600,40.21) (700,39.66) (800,40.09) (900,40.67) (1000,41.16)};
\addplot[cUnif!35, thin, mark=none, forget plot] coordinates {(250,40.19) (300,40.33) (400,40.41) (500,40.55) (600,40.52) (700,40.33) (800,40.81) (900,40.71) (1000,41.05)};
\addplot[cUnif, thick, mark=*, mark size=1.8pt] coordinates {(250,40.24) (300,40.35) (400,40.09) (500,40.09) (600,40.37) (700,39.99) (800,40.45) (900,40.69) (1000,41.11)};
\legend{{TSC, mean of 2 seeds}, {uniform, mean of 2 seeds}}
\end{axis}
\end{tikzpicture}
\caption{\textbf{Accuracy from 250 to 1{,}000 steps at $8{\times}8$} (Qwen3-1.7B, each method at its
best learning rate, scored every 100 steps). Thin lines are the individual seeds.}
\label{fig:durability}
\end{figure}

We extended the two configurations with the largest gains to 1{,}000 steps and scored both methods every 100
steps (Figure~\ref{fig:durability}). Scores of a single run change by two to three points
from one checkpoint to the next, so we compare averages over the checkpoints from step 600
to step 1{,}000.

At $\shape{8}{8}$, TSC averages 41.85\% and uniform sampling 40.52\%. Each of the two TSC
runs (42.50\% and 41.19\%) is above each uniform run (40.36\% and 40.69\%). As at 250 steps,
the gain is in reliability. Over the same checkpoints, pass$^{16}$ is 15.9\% for TSC and
14.1\% for uniform sampling, and pass@16 is the same for both (74.9\% and 75.0\%). At
$\shape{32}{4}$, TSC averages 42.23\% and uniform sampling 41.16\%, and each TSC run (42.05\%
and 42.42\%) is again above each uniform run (41.39\% and 40.92\%). Here, pass$^{16}$ is 1.48
points higher for TSC and pass@16 is 0.54 points higher.

Longer training does not lower the share of zero-variance groups under uniform sampling. Its share of
zero-variance groups at $\shape{8}{8}$ stays at about half in every block of 250 steps. The share for TSC rises slowly, from 10.7\% in the first block to 14.2\% in the last.

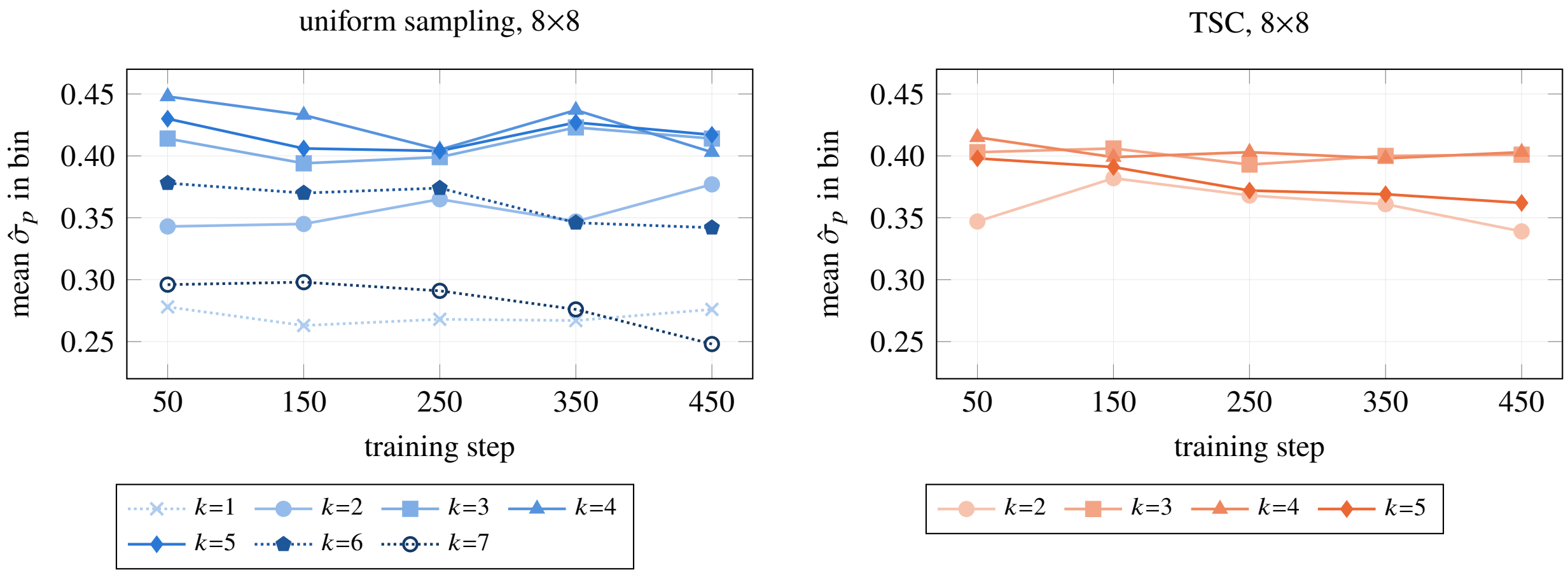
\begin{figure}[tbp]
\centering
\colorlet{uK1}{cUnif!38}\colorlet{uK2}{cUnif!50}\colorlet{uK3}{cUnif!60}\colorlet{uK4}{cUnif!80}\colorlet{uK5}{cUnif}\colorlet{uK6}{cUnif!72!black}\colorlet{uK7}{cUnif!48!black}
\colorlet{tK2}{cOurs!40}\colorlet{tK3}{cOurs!60}\colorlet{tK4}{cOurs!80}\colorlet{tK5}{cOurs}
\begin{minipage}[t]{.49\linewidth}\centering
\begin{tikzpicture}
\begin{axis}[
  width=\linewidth, height=4.8cm, title={uniform sampling, $8{\times}8$}, title style={font=\small},
  xlabel={training step}, ylabel={mean $\hat\sigma_p$ in bin},
  xmin=20, xmax=480, ymin=0.22, ymax=0.47, ytick={0.25,0.30,0.35,0.40,0.45}, yticklabel style={/pgf/number format/fixed, /pgf/number format/fixed zerofill, /pgf/number format/precision=2}, xtick={50,150,250,350,450},
  tick label style={font=\small}, label style={font=\small},
  legend to name=binspreadlegendU, legend columns=4, legend cell align=left,
  legend style={font=\scriptsize, /tikz/every even column/.append style={column sep=3pt}},
  grid=major, grid style={line width=.1pt, draw=gray!18},
]
\addplot[uK1, thick, densely dotted, mark=x, mark size=2.6pt, mark options={solid, thick}] coordinates {(50,0.278) (150,0.263) (250,0.268) (350,0.267) (450,0.276)};
\addplot[uK2, thick, mark=*] coordinates {(50,0.343) (150,0.345) (250,0.365) (350,0.347) (450,0.377)};
\addplot[uK3, thick, mark=square*] coordinates {(50,0.414) (150,0.394) (250,0.399) (350,0.423) (450,0.414)};
\addplot[uK4, thick, mark=triangle*] coordinates {(50,0.448) (150,0.433) (250,0.405) (350,0.437) (450,0.403)};
\addplot[uK5, thick, mark=diamond*] coordinates {(50,0.430) (150,0.406) (250,0.404) (350,0.427) (450,0.417)};
\addplot[uK6, thick, densely dotted, mark=pentagon*, mark options={solid}] coordinates {(50,0.378) (150,0.370) (250,0.374) (350,0.346) (450,0.342)};
\addplot[uK7, thick, densely dotted, mark=o, mark options={solid}] coordinates {(50,0.296) (150,0.298) (250,0.291) (350,0.276) (450,0.248)};
\legend{$k{=}1$,$k{=}2$,$k{=}3$,$k{=}4$,$k{=}5$,$k{=}6$,$k{=}7$}
\end{axis}
\end{tikzpicture}\\[2pt]
\pgfplotslegendfromname{binspreadlegendU}
\end{minipage}\hfill
\begin{minipage}[t]{.49\linewidth}\centering
\begin{tikzpicture}
\begin{axis}[
  width=\linewidth, height=4.8cm, title={TSC, $8{\times}8$}, title style={font=\small},
  xlabel={training step}, ylabel={mean $\hat\sigma_p$ in bin},
  xmin=20, xmax=480, ymin=0.22, ymax=0.47, ytick={0.25,0.30,0.35,0.40,0.45}, yticklabel style={/pgf/number format/fixed, /pgf/number format/fixed zerofill, /pgf/number format/precision=2}, xtick={50,150,250,350,450},
  tick label style={font=\small}, label style={font=\small},
  legend to name=binspreadlegendT, legend columns=4, legend cell align=left,
  legend style={font=\scriptsize, /tikz/every even column/.append style={column sep=3pt}},
  grid=major, grid style={line width=.1pt, draw=gray!18},
]
\addplot[tK2, thick, mark=*] coordinates {(50,0.347) (150,0.382) (250,0.368) (350,0.361) (450,0.339)};
\addplot[tK3, thick, mark=square*] coordinates {(50,0.403) (150,0.406) (250,0.393) (350,0.400) (450,0.401)};
\addplot[tK4, thick, mark=triangle*] coordinates {(50,0.415) (150,0.399) (250,0.403) (350,0.398) (450,0.403)};
\addplot[tK5, thick, mark=diamond*] coordinates {(50,0.398) (150,0.391) (250,0.372) (350,0.369) (450,0.362)};
\legend{$k{=}2$,$k{=}3$,$k{=}4$,$k{=}5$}
\end{axis}
\end{tikzpicture}\\[2pt]
\pgfplotslegendfromname{binspreadlegendT}
\end{minipage}
\caption{\textbf{Reward spread within each solve-count bin over 500 steps} (Qwen3-1.7B,
$\shape{8}{8}$, each method at its best learning rate, two seeds pooled, means over blocks of 100
steps). $\hat\sigma_p=\sqrt{\hat\mu(1-\hat\mu)}$ is computed from the eight rewards of each served
group, where $\hat\mu$ is the mean reward of the group. For uniform sampling, the bins $k=1$ to 7 are shown. The two end bins, $k=0$ and $k=8$,
have a mean spread of about 0.1 under uniform sampling and are not shown. For TSC, only the
bins it serves at least 50 times per block are shown ($k=2$ to 5).
Uniform sampling is drawn in shades of blue and TSC in shades of orange. Darker lines are bins
with a higher solve count $k$. Dotted lines mark the three bins that are not shown for TSC
($k=1$, $k=6$ and $k=7$).}
\label{fig:binspread}
\end{figure}

\paragraph{Reward spread within each bin.} TSC concentrates on a few bins, so the reward spread in those bins could fall as the model
learns their prompts. The 500-step runs show no sign of this
(Figure~\ref{fig:binspread}). Under TSC at $\shape{8}{8}$, the two bins that it serves most,
$k=3$ and $k=4$, keep a mean reward spread of about 0.40 for all 500 steps. Under uniform sampling, the spread is stable in most bins. The easiest bin with mixed
rewards, $k=7$, loses spread as the model starts to solve its prompts every time (from 0.30
to 0.25), and the hard bin $k=2$ gains a little (from 0.34 to 0.38).

\subsection{Interventions on Within-Step Entropy}\label{app:causal}\label{sec:causal}

If within-step entropy caused the gain, raising it inside one configuration should raise the gain. We tested this in two ways.

\paragraph{Temperature on the Thompson draw.} A temperature above one flattens the
distribution over bins, and a temperature below one sharpens it. We ran temperatures of 0.5,
1, 2 and 4 at $\shape{4}{4}$ and $\shape{8}{8}$ with two seeds each. The entropy moves from
0.53 to 1.53 bits at $\shape{4}{4}$ and from 0.55 to 2.05 bits at $\shape{8}{8}$. The gain
over uniform sampling does not follow the entropy. At $\shape{8}{8}$, the gain is $+1.35$, $+2.01$,
$+0.47$ and $+1.29$ for the four temperatures, and at $\shape{4}{4}$, it is $-1.42$, $-1.36$,
$-2.48$ and $-1.78$. The sign is set by the configuration, and the default temperature of 1
is the best in both.

\paragraph{Stratified draws.} A temperature also changes which bins are served. The second
test changes only how the prompts of one step are spread over the bins. We ran it at
$\shape{4}{4}$, the configuration where TSC is furthest behind uniform sampling, and asked
whether a stratified draw improves TSC there. If TSC selects bin $c$ with probability
$q_c$, the stratified draw gives this bin either $\lfloor Pq_c\rfloor$ or $\lceil Pq_c\rceil$
of the $P$ prompts of the step, where independent draws could give it all or none of them. We
ran two variants with two seeds each. One takes $q$ from the current beliefs, and the other
from an average of $q$ over about the last 20 steps. Entropy rises from 0.97 to 1.37 bits,
which is above the 1.14 bits where the fitted line crosses zero, and the share of
zero-variance groups stays the same (Table~\ref{tab:n4}). Averaged over the four stratified
runs, accuracy rises by 0.89 points over standard TSC, which closes about two thirds of the
gap to uniform sampling. With $q$ from the current beliefs, the gap falls from 1.36 points to
0.35. The relationship across configurations predicts a larger increase, of about 2.0 points.

The stratified draw therefore improves TSC in a configuration with few prompts per step.
Inside a fixed configuration, the effect of entropy is smaller than the relationship across
configurations suggests.

\begin{table}[tbp]
\centering\small
\begin{tabular}{lcccc}
\toprule
sampler ($4{\times}4$, rate $10^{-5}$) & entropy (bits) & zero-variance & avg@16 (seeds) & gain over uniform \\
\midrule
uniform & 1.51 & 61\% & 40.95 (41.65 / 40.25) & --- \\
TSC & 0.97 & 28\% & 39.59 (40.02 / 39.15) & $-1.36$ \\
TSC, stratified (averaged $q$) & 1.37 & 28\% & 40.37 (41.35 / 39.38) & $-0.58$ \\
TSC, stratified (current $q$) & 1.37 & 29\% & 40.60 (40.85 / 40.35) & $-0.35$ \\
\bottomrule
\end{tabular}
\caption{\textbf{Stratified draws at the $4{\times}4$ configuration} (1.7B, 250 steps,
two seeds).}
\label{tab:n4}
\end{table}

\subsection{Transfer to Other Benchmarks}\label{sec:transfer}

We scored the two configurations with the largest gains, both methods and both seeds, on MATH500 and on the 2024
and 2025 AIME exams (Table~\ref{tab:transfer}). On MATH500, TSC is ahead by $+0.9$ points at
$\shape{8}{8}$ and by $+1.0$ at $\shape{32}{4}$, about half of the gain on the held-out set.
Pooled over the two configurations, the gain at step 250 is $+0.93$ \ci{+0.42}{+1.42} on
MATH500, $+3.23$ \ci{+1.15}{+5.52} on AIME25 and $+0.52$ \ci{-1.15}{+2.14} on AIME24. The AIME
exams have 30 problems each, so their intervals are wide. As on the held-out set, pass@16 on
MATH500 is unchanged. At 500 steps, the pooled MATH500 gain is $+0.5$ \ci{-0.1}{+1.1}.

\begin{table}[tbp]
\centering\small
\begin{tabular}{llrrrrr}
\toprule
$P{\times}G$ & benchmark & untrained & uniform & TSC & gain [95\% CI] & gain at 500 steps\\
\midrule
$\shape{8}{8}$  & MATH500 & 58.1 & 59.2 & 60.1 & $+0.9$ \ci{+0.1}{+1.7} & $-0.3$ \ci{-1.1}{+0.6}\\
      & AIME24  & 11.9 & 13.0 & 14.6 & $+1.6$ \ci{-0.5}{+3.9} & $+1.9$\\
      & AIME25  &  9.6 & 10.3 & 12.9 & $+2.6$ \ci{-0.1}{+5.9} & $-0.5$\\
\midrule
$\shape{32}{4}$ & MATH500 & 58.1 & 59.6 & 60.6 & $+1.0$ \ci{+0.3}{+1.6} & $+1.3$ \ci{+0.4}{+2.2}\\
      & AIME24  & 11.9 & 14.7 & 14.2 & $-0.5$ \ci{-2.8}{+1.8} & $+0.7$\\
      & AIME25  &  9.6 & 11.2 & 15.0 & $+3.9$ \ci{+0.8}{+7.2} & $+2.8$\\
\bottomrule
\end{tabular}
\caption{\textbf{Transfer to external benchmarks} (avg@16, two seeds per method, each method at its
own best learning rate, 250 steps unless noted). Intervals are paired bootstraps over
problems.}
\label{tab:transfer}
\end{table}

\subsection{LoRA Rank}\label{app:rank}

All main runs use LoRA rank 64. To check whether the gain depends on the capacity of the
adapter, we repeated $\shape{8}{8}$ with rank 128 ($\alpha=256$, so $\alpha/r$ is unchanged).
We ran both methods at five learning rates with one seed, and a second seed at the best rate
of each method (Table~\ref{tab:rank}).

\begin{table}[tbp]
\centering\small
\begin{tabular}{lccccc}
\toprule
learning rate & $1.25{\times}10^{-6}$ & $2.5{\times}10^{-6}$ & $5{\times}10^{-6}$ & $10^{-5}$ & $2{\times}10^{-5}$\\
\midrule
uniform (seed 1) & 39.80 & \textbf{40.10} & 38.73 & 38.15 & 39.67\\
TSC (seed 1) & 40.04 & \textbf{40.90} & 40.65 & 40.20 & 40.17\\
gain & $+0.24$ & $+0.80$ & $+1.92$ & $+2.05$ & $+0.50$\\
\midrule
\multicolumn{6}{l}{two seeds at the best rate ($2.5\times10^{-6}$): uniform 40.20, TSC 40.68, gain $+0.48$}\\
\bottomrule
\end{tabular}
\caption{\textbf{LoRA rank 128 at $8{\times}8$} (Qwen3-1.7B, held-out avg@16 at step 250). At rank 64,
the two-seed scores at the best rates are 40.24 for uniform sampling and 42.25 for TSC.}
\label{tab:rank}
\end{table}

Uniform sampling is unchanged, with 40.20\% against 40.24\% at rank 64. The best score of TSC
falls from 42.25\% to 40.68\%, so the gain at the best rates falls from $+2.01$ to $+0.48$.
TSC is still above uniform sampling at all five learning rates, by $+1.10$ on average, and
both TSC seeds at the best rate are above both uniform seeds. At rank 64, the best rate for TSC was $2\times10^{-5}$, and at rank 128, it is $2.5\times10^{-6}$. Part of the gain at rank
64 therefore comes from TSC tolerating a higher learning rate at that rank, and the size of
the gain should not be expected to carry over to other adapters.

\subsection{Experiments on Qwen3-4B}\label{app:4b}

All 4B experiments use eight rollouts per prompt, one seed per method and bins measured
with the 4B model. All runs use eight prompts per step, except the 64-prompt runs below.

\paragraph{TSC and uniform sampling.} A matched pair with the same learning rate
($10^{-5}$) and a checkpoint every 25 steps gives a gain of $+0.69$ points at step 250 and
$+0.49$ on average over twelve checkpoints up to step 300. The gain is positive at nine of
the twelve checkpoints, and scores vary by about one point between checkpoints. The same
configuration and rate give $+1.59$ on 1.7B.

\paragraph{DAPO.} The comparison is at step 500, where TSC and DAPO have both trained on
32{,}000 rollouts. TSC scores 57.6\% and DAPO scores 57.5\%, a paired difference (TSC minus DAPO) of $+0.14$
points \ci{-1.03}{+1.30}. Up to that step, TSC generated 32{,}000 rollouts and DAPO generated
72{,}984, which is 2.28 times as many (Table~\ref{tab:4b}). The earlier DAPO
checkpoints, at steps 100 to 400, score between 56.1\% and 57.4\%.

\paragraph{Larger batches.} With 64 prompts per step, uniform sampling generates eight times as
many rollouts as TSC with eight prompts per step and reaches the same accuracy at step 500
(Table~\ref{tab:4b}). Its best checkpoint, 58.2\% at step 300, comes after 153{,}600 rollouts.
TSC with 64 prompts per step finishes level with uniform sampling at that batch (57.57\%
against 57.55\%), although it removes most of the zero-variance groups. Both 64-prompt runs
use a learning rate of $3\times10^{-5}$.

\begin{table}[tbp]
\centering\small
\begin{tabular}{lcccc}
\toprule
method & prompts per step & rollouts generated & zero-variance groups & accuracy\\
\midrule
uniform & 8 & 32{,}000 & 52\% & 55.73\\
TSC & 8 & 32{,}000 & 12\% & \textbf{57.60}\\
DAPO & 8 & 72{,}984 & discarded & 57.46\\
uniform & 64 & 256{,}000 & 54\% & 57.55\\
TSC & 64 & 256{,}000 & 14\% & 57.57\\
\bottomrule
\end{tabular}
\caption{\textbf{Larger batches and DAPO on Qwen3-4B} (one seed per method, held-out avg@16
at step 500). Rollouts generated counts every rollout up to step 500. The zero-variance column
is the share of generated groups whose rewards are all equal. DAPO discards these groups and
generates more until its batch is full.}
\label{tab:4b}
\end{table}

\FloatBarrier
\section{Proofs and Identities}\label{app:proofs}

\paragraph{Notation.} Consider one prompt that the current model solves with probability $p$.
Its reward is 1 with probability $p$ and 0 otherwise, so the reward has mean $p$ and variance
$\sigma^2=p(1-p)$. A group of $G$ rollouts of this prompt contains $k\sim\mathrm{Binom}(G,p)$
correct rollouts and has mean reward $\bar r=k/G$. The $k$ correct rollouts each differ from
$\bar r$ by $(G-k)/G$, and the $G-k$ incorrect rollouts by $k/G$, so the squared deviations sum
to $k(G-k)/G$. Two standard deviations of the $G$ rewards appear in this paper. The reward
spread of Section~\ref{sec:identity} divides this sum by $G$,
\[
\sigp^2=\frac{k(G-k)}{G^2},
\]
and the sample standard deviation $s$ in Equation~\eqref{eq:grpo} divides it by $G-1$,
\[
s^2=\frac{k(G-k)}{G(G-1)}=\frac{G}{G-1}\,\sigp^2 .
\]
TSC uses the reward spread of each served group as its observation of $\sigma$. The main text
writes $\hat\sigma_p$ when it refers to this observation; it is the same quantity as $\sigp$.

\paragraph{Proof of Proposition~\ref{prop:identity}.} The absolute deviations of the $G$
rewards from $\bar r$ sum to $k\frac{G-k}{G}+(G-k)\frac kG=\frac{2k(G-k)}{G}$.
Equation~\eqref{eq:grpo} divides each deviation by $s$, so
\[
\begin{aligned}
\sum_i|A_i|&=\frac{2k(G-k)/G}{s}=2\sqrt{k(G-k)}\sqrt{\tfrac{G-1}{G}}\\
&=2\sqrt{G(G-1)}\,\frac{\sqrt{k(G-k)}}{G}=2\sqrt{G(G-1)}\,\sigp .
\end{aligned}
\]

\paragraph{How the advantage is split.} Away from $k=G/2$, the advantage is split unevenly
across rollouts. At $G=8$ and $k=1$, the single correct rollout receives $A=+2.475$ and each of
the seven incorrect rollouts receives $-0.354$. This is one reason to bin by $k$ rather than by
$\sigp$, because $\sigp(k)=\sigp(G-k)$ would put nearly unsolvable and nearly solved prompts in
the same bin.

\paragraph{Mean and spread.} Under a binary reward, $\sigp^2=k(G-k)/G^2=\bar r(1-\bar r)$
exactly. The spread therefore adds no information to the mean, and the solve count is the
finest partition that the signal supports. This does not hold for graded rewards
(Appendix~\ref{app:betabin}).

\paragraph{Population version.} Under binary rewards, the GRPO advantages have a closed form
in terms of $p$ when the mean and standard deviation of the policy replace those of the group
\citep{mroueh2025grpo}. A correct answer, which occurs with probability $p$, receives
$A_+=(1-p)/\sigma$, and an incorrect answer, which occurs with probability $1-p$, receives
$A_-=-p/\sigma$. The expected magnitude of one advantage is therefore
\[
\mathbb E|A| \;=\; p\,\frac{1-p}{\sigma}+(1-p)\,\frac{p}{\sigma}
\;=\;\frac{2p(1-p)}{\sigma}\;=\;2\sigma ,
\]
and a group of $G$ rollouts carries $2G\sigma$ in expectation. Proposition~\ref{prop:identity}
is the exact version of this result for one group. Its constant is $2\sqrt{G(G-1)}$ and not
$2G$ because Equation~\eqref{eq:grpo} uses the sample standard deviation $s$, which divides by
$G-1$.

\paragraph{An upper bound on the gain from selection.} Selecting prompts by bin can raise the advantage magnitude per rollout by a factor of at most 2.15 over
uniform sampling. We compute this bound from the mean of $|A_i|$ over the eight rollouts of a
group, where $A_i$ is the advantage of rollout $i$ (Equation~\eqref{eq:grpo}). By
Proposition~\ref{prop:identity}, this mean is proportional to the reward spread of the group.
For each bin, we pick a prompt from the bin at random and generate a new group of eight
rollouts for it. We compute the expected mean $|A_i|$ of this group, averaged over the prompts
of the bin and over the outcomes of the eight rollouts. A bin contains prompts with different
solve probabilities. For example, the bin with no correct rollouts contains prompts that the
model cannot solve and prompts that it solves now and then but failed eight times by chance.
We first estimate how common each solve probability is in the whole pool, by choosing the
shares that best reproduce the pool's distribution of solve counts. To fit these shares, we
restrict solve probabilities to a grid of 201 values between 0 and 1, start from equal shares,
and run 400 iterations of the EM algorithm. Solve counts from eight rollouts do not determine
these shares uniquely, so this fit is one plausible choice. From these shares, we compute which
solve probabilities the prompts in each bin have, and we average the expected reward spread of
a new group of eight rollouts over them. The bins with no correct rollouts and with eight
correct rollouts then have an expected reward spread of about 0.06 and not zero. With each bin
weighted by its share of the pool measured with Qwen3-4B, uniform sampling gives an expected mean $|A_i|$ of 0.372.
Serving only the bins with three to five correct rollouts gives 0.782, 2.10 times as much, and
serving only the bin with four correct rollouts gives 0.800, 2.15 times as much. This bin has
the highest value of the nine, so no mix of bins can give more.

\section{The Beta--Binomial Model and Graded Rewards}\label{app:betabin}

\paragraph{Fit.} We model the solve rate $p$ of a prompt as a draw from a
$\mathrm{Beta}(a,b)$ distribution. Maximum likelihood on the nine bin counts measured with
Qwen3-4B gives $a=0.3231$ and $b=0.3045$. Both are below one, which produces a U-shaped
histogram, and the fitted and observed bin counts agree to within 12\% in every bin. The
share of zero-variance groups with $G$ rollouts is
$\Pr[\text{all correct}]+\Pr[\text{all wrong}]=[B(a{+}G,b)+B(a,b{+}G)]/B(a,b)$. Table~\ref{tab:betabin}
compares the shares predicted by this fit with the shares measured during training of
Qwen3-1.7B. The predicted shares are 3 to 5 points above the measured ones. The fit uses the
bin counts of Qwen3-4B, and the measurements come from Qwen3-1.7B. The fit reproduces the
slow decline with $G$.

\begin{table}[tbp]
\centering\small
\begin{tabular}{lcccccc}
\toprule
$G$ & 4 & 8 & 16 & 32 & 64 & 256\\
\midrule
predicted share of zero-variance groups & .650 & .523 & .420 & .338 & .272 & .176\\
predicted share of groups with all rollouts correct & .339 & .274 & .222 & .180 & .146 & .095\\
measured share of zero-variance groups (uniform sampling, Qwen3-1.7B) & .618 & .475 & .371 & .307 & --- & ---\\
\bottomrule
\end{tabular}
\caption{\textbf{Share of zero-variance groups, predicted by the Beta--Binomial fit and
measured in training} ($G$ is the number of rollouts per prompt). The first two rows are
predictions of the Beta--Binomial distribution fitted to the bin counts of Qwen3-4B. The
second row is the part of the first row in which all $G$ rollouts are correct. The third row
is measured during training of Qwen3-1.7B with uniform sampling.}
\label{tab:betabin}
\end{table}

\paragraph{Graded rewards.} We model a graded reward with $K$ criteria as
$r=\mathrm{Binom}(K,p_c)/K$. The success probability per criterion is $p_c=a_f+(1-2a_f)p$,
where $a_f$ is a floor of partial credit that even a poor answer receives, and $p$ follows
the fitted Beta distribution. With $G=8$, $K=10$ and a floor of 0.15, the share of
zero-variance groups falls almost to zero. The difference in $\sigp$ between prompts falls as
well, and its coefficient of variation drops from 0.67 to 0.13. The floor causes this drop, not
$K$: with $a_f=0$, the coefficient of variation is unchanged. A selector based on
variance relies on this difference. Measuring it on a task with graded rewards can help to
judge how useful such a selector will be there.

\section{The Kalman Sampler and the Choice of Bins}\label{app:kalman}

\paragraph{Update.} The served bin is updated by a scalar Kalman step with observation
$z=\hat\sigma_p$: $v^{\text{pred}}=v_c+\tau$, $g=v^{\text{pred}}/(v^{\text{pred}}+R)$,
$m_c\leftarrow m_c+g(z-m_c)$ and $v_c\leftarrow(1-g)v^{\text{pred}}$. The variance of every
other bin grows as $v_{c'}\leftarrow\min(v_{c'}+\tau,v_0)$ once per optimizer step.

\paragraph{Constants.} We use an observation noise of $R=0.0356$, which is the measured
variance of $\hat\sigma_p$ within a bin, a prior mean of 0.3, a prior variance of
$v_0=0.01$ and a variance growth of $\tau=10^{-5}$. The growth $\tau$ balances two errors.
When the true spread of a bin drifts by $d$ per update, the estimate lags behind by about
$d/g$ and has a noise variance of about $gR/2$, where the steady-state gain is
$g\approx\sqrt{\tau/R}$. Minimizing the sum of the squared lag and the noise variance gives $\tau^\star=(4d^2)^{2/3}R^{1/3}$. The drift
we measured is $6\times10^{-5}$ to $2\times10^{-4}$ per update, which puts $\tau^\star$
between $2\times10^{-6}$ and $10^{-5}$. The prior variance must be small compared with the
squared differences between the bin means. Otherwise, the Thompson draws of bins that have
not been served recently are dominated by noise, and the sampler behaves like uniform
sampling.

\paragraph{Bins, individual prompts and topics.} At our budgets, a prompt is served 0.03 to
0.21 times per run on average, so an estimate for each prompt would stay close to its prior.
Table~\ref{tab:partition} compares ways of grouping the prompts, measured on a
uniform-sampling run of Qwen3-4B. The intraclass correlation (ICC) measures how well the
group of a prompt predicts its reward spread $\hat\sigma_p$. It is the variance of
$\hat\sigma_p$ between the groups divided by the total variance of $\hat\sigma_p$ over all
prompts. An ICC of 1 means that all prompts in a group have the same spread, and an ICC of 0
means that all groups have the same mean spread. The ICC is 0.42 for the solve-count bins and
0.11 for 21 clusters defined by topic and a heuristic difficulty level. The rank reliability of the
solve-count bins is 0.994 (Table~\ref{tab:partition}), so their ranking is estimated almost
without error.

\begin{table}[tbp]
\centering\small
\begin{tabular}{lccc}
\toprule
partition (4B pool) & bins & ICC & rank reliability\\
\midrule
solve-count bins & 9 & \textbf{0.42} & \textbf{0.994}\\
topic $\times$ heuristic difficulty & 21 & 0.11 & 0.958\\
per prompt, 1 visit & 37{,}970 & --- & 0.634\\
per prompt, 10 visits & 37{,}970 & --- & 0.945\\
\bottomrule
\end{tabular}
\caption{\textbf{Ways of grouping the prompts} (uniform-sampling run of Qwen3-4B). Each row is
one way of sorting the prompts into bins. The bins are the solve-count bins in the first row,
21 clusters of topic and difficulty in the second row, and single prompts in the last two
rows. ICC is the intraclass correlation, the variance of $\hat\sigma_p$ between bins divided
by its total variance over all prompts. Rank reliability is the share of the variance in the
estimated bin means that comes from true differences between bins, with the rest coming from
sampling noise.}
\label{tab:partition}
\end{table}

\paragraph{Noisy bin labels.} Eight rollouts give a noisy solve count. When we measure 500
held-out prompts again with sixteen rollouts, 34\% of them move to a different bin. This
matters little for selection, because the sampler needs the mean spread of each bin and not
the true bin of each prompt. Binning the 500 prompts with the finer measurement changes no
bin mean by more than 0.025 and preserves the ranking of the bins (Spearman $\rho=0.95$).

\section{Implementation of the Compared Methods}\label{app:comparators}

\paragraph{Static headroom.} The distribution over bins is fixed before training, and the
bin of each prompt in a step is drawn from it independently, with probability proportional to
the bin's weight. A bin of size $|c|$
and solve count $k$ has weight $|c|\cdot(1[k\ge1]-k/8)$. On the 1.7B pool, 38\% of the draws
come from the one-of-eight bin, 23\% from the two-of-eight bin and none from the never-solved
and always-solved bins. This method therefore serves harder prompts than TSC, which
concentrates on the bins with three to five of eight rollouts solved.

\paragraph{DAPO.} We add DAPO's dynamic sampling to the same trainer and recipe as the other
methods. In each step, it generates groups for uniformly sampled prompts, keeps the groups
with nonzero variance and repeats for up to eight rounds until the batch is full. Each round
requests 1.6 times the number of groups still missing. The generation cost is the number of
rollouts recorded in its log. VCRL \citep{jiang2025vcrl} also scores prompts only after their
rollouts are generated, so it pays for the groups it discards, as DAPO does. It refills the
batch from a replay bank instead of generating new rounds, and its paper does not specify the
generation cost of the replayed prompts. We did not run it separately.

\paragraph{SPaCe.} We follow the scheduler described in the SPaCe paper
\citep{do2025sparft}. At the start of each step, it draws
$\tilde\mu_k\sim\mathcal N(m_k,1/(n_k+\epsilon))$ for every cluster $k$, where
$m_k=-R_k/(n_k+\epsilon)$, $R_k$ is the sum of the cluster's mean rewards over the steps in
which it was served and $n_k$ is the number of those steps. The cluster with the largest
draw is served for the whole step. A cluster stagnates when, in $T=10$ consecutive steps in which it is served, its mean
reward fails to exceed its mean reward in the previous such step by at least $\delta$. The
value of a stagnating cluster is reduced by $\gamma$ until its mean reward improves again. We
use $\epsilon=10^{-6}$ and $T=10$ as in the paper. The paper does not report $\gamma$ or
$\delta$, and we set $\gamma=1$ and $\delta=0.01$. Because rewards lie in $[0,1]$, $\gamma=1$
places a stagnating cluster below every cluster that is still improving. The SPaCe paper
describes $\gamma$ as a small constant. In a simulation with the solve rates of our clusters,
raising $\gamma$ from 0 to 1 moves SPaCe away from the never-solved clusters and lowers its
share of zero-variance groups from 70\% to 47\%, so $\gamma=1$ is the setting most favorable
to SPaCe here.
We run the scheduler on the full pool and do not reduce the training set to about one
hundred examples as the SPaCe paper does, because our question is which selection rule to
use on the full pool.

\paragraph{Clusters for the SPaCe comparison.} The topic is one of the seven clusters of
sentence embeddings that come with the dataset. The difficulty has three levels by the solve
count of the 1.7B model: never solved (34.3\% of the pool), solved one to four times
(33.0\%) and solved five to eight times (32.7\%). The 21 clusters contain between 691 and
3{,}044 prompts each. TSC runs on these clusters with unchanged constants.

\section{The Code Task}\label{app:code}

\paragraph{Data.} From KodCode-Light-RL-10K, we set aside 500 problems, stratified by the
difficulty label of the dataset, as a held-out set. We then kept only problems whose
reference solution passes its own tests in our sandbox. This leaves 9{,}225 of the 9{,}500
training problems and 483 of the 500 held-out problems. The held-out set is used to choose
the learning rate of each method. HumanEval+ (164 problems) and MBPP+ (378 problems) are
used only as test sets. The training reward runs the same test harness as the evaluation.

\paragraph{Bins.} One pass of the untrained model with eight rollouts per problem gives each
problem a solve count, and the nine counts are the bins. The model never solves 1{,}631
problems and always solves 4{,}632, which leaves 2{,}962 problems in between. The pass
generates 73{,}800 rollouts.

\paragraph{TSC constants.} The constants were set on math, so we checked them on code before
training. The mean $\sigp$ per bin ranges from 0.03 and 0.06 in the two extreme bins to 0.42
in the bins with three to five solved rollouts. The variance of $\sigp$ within a bin is 0.018
to 0.024, below the value $R=0.0356$ measured on math. In a simulation of the
sampler with these values, the share of zero-variance groups falls from 68\% to between 5\%
and 8\% for the math constants and for every other setting in a sweep around them, so we
kept the math constants.

\paragraph{Training.} Training uses the math recipe without change. To choose the learning
rate, we ran each method for 250 steps at $2.5\times10^{-6}$, $10^{-5}$ and $2\times10^{-5}$ in
both configurations with one seed, and with a second seed at the rate that is best on the
held-out set. This rate is $10^{-5}$ in all four cases. We then trained a separate set of
runs at $10^{-5}$, with two seeds per method and a checkpoint every 50 steps. These runs last
500 steps at $\shape{8}{8}$ and 250 steps at $\shape{32}{4}$, so that both configurations
reach 32{,}000 rollouts. All results in Section~\ref{sec:code} and Table~\ref{tab:code} come
from these runs.

\paragraph{Test-set formats.} Every held-out KodCode prompt and every HumanEval+ prompt
contains the signature of the function to write (483 of 483 and 164 of 164). No MBPP+ prompt
does. An MBPP+ prompt gives a one-sentence task and one example test, from which the model
must infer the function name and arguments.

\begin{table}[tbp]
\centering\small
\begin{tabular}{lcccccccc}
\toprule
 & \multicolumn{2}{c}{KodCode held-out} & \multicolumn{2}{c}{HumanEval+} & \multicolumn{2}{c}{MBPP+} & \multicolumn{2}{c}{zero-variance groups}\\
$P{\times}G$ (steps) & uniform & TSC & uniform & TSC & uniform & TSC & uniform & TSC\\
\midrule
$\shape{8}{8}$ (500) & 70.60 & \textbf{71.86} & 66.22 & \textbf{67.25} & \textbf{54.87} & 53.35 & 63\% & 23\% \\
$\shape{32}{4}$ (250) & 69.75 & \textbf{70.50} & 66.33 & \textbf{66.73} & \textbf{54.47} & 53.88 & 72\% & 36\% \\
\bottomrule
\end{tabular}
\caption{\textbf{Code generation with unit tests after 32{,}000 training rollouts} (Qwen3-1.7B,
avg@16, mean of two seeds per method at learning rate $10^{-5}$). The share of zero-variance
groups is averaged over all training steps. The higher score of each pair is in bold. The
untrained model scores 66.08 on the held-out set, 63.95 on HumanEval+ and 53.41 on MBPP+.}
\label{tab:code}
\end{table}

\section{The Agent Task}\label{app:agent}

\paragraph{Agent and reward.} The policy is Nemotron 3.5 Lightning (30B parameters, 3B
active), trained with LoRA (rank 64). It
uses the tools of Samaya's research agent, one of the systems evaluated in
FrontierFinance~\citep{zhang2026frontierfinance}. The rubrics are written by people, in the
same form as the benchmark's rubrics, and language-model judges decide which criteria a
report meets. We report the mean fraction of criteria met.

\paragraph{Starting point and baseline.} Every run starts from the same checkpoint, the
result of the first three stages of the Sterling recipe \citep{koyluoglu2026sterling}: two
stages of SFT on teacher traces and one stage of offline GRPO on eight teacher traces. It
scores 59.9\% on the held-out questions. The uniform-sampling baseline is the first 22 steps
of the fourth stage, online GRPO. The uniform-sampling baseline and the TSC run use the same
recipe, with 16 questions and 8 rollouts per step, a learning rate of $7\times10^{-5}$, a KL
penalty of 0.005 to the starting checkpoint, and advantages centered on the group mean
without division by the standard deviation. The importance ratio of each token is capped at 2, as in
CISPO \citep{minimax2025m1}, and the loss is averaged at the prompt level, so that each
question contributes equally \citep{khatri2025scalerl}.

\paragraph{Online sampler.} Each question has an estimate $\hat p$ of its mean reward. The
estimate starts from older rollouts of the same question and is updated after each group as
$\hat p\leftarrow\tfrac12\hat p+\tfrac12\bar r$, where $\bar r$ is the mean reward of the
group. Before each step, the questions are divided into five bins of equal size by $\hat p$,
numbered $b=0$ to 4 from the lowest mean reward to the highest. Each bin keeps a Gaussian
belief over the reward spread of its groups, updated by the Kalman step of
Algorithm~\ref{alg:tsc} with constants measured on the stored rollouts ($R=4\times10^{-3}$,
$v_0=5\times10^{-4}$, $\tau=10^{-6}$ and a prior mean of 0.112). The sampler draws
$\tilde u_b$ from the belief of each bin, picks bin $b$ with probability proportional to
$\max(\tilde u_b,10^{-3})\,e^{-|b-2|}$, and serves the question in that bin that has been
served least.

\section{Training Recipe and Reproducibility}\label{app:repro}

\paragraph{Model and training.} We use Qwen3-1.7B, Qwen3-0.6B and Qwen3-4B in non-thinking
mode, fine-tuned with LoRA (rank 64, $\alpha=128$, dropout 0.05). GRPO uses group-normalized
advantages, a KL penalty of $\beta=0.01$ to the reference model and a clip range of 0.2. The
implementation uses TRL \citep{vonwerra2020trl} for training and vLLM \citep{kwon2023vllm}
for generation. Training rollouts are generated at temperature 1.0 with a budget of 4{,}096
tokens. Learning rates range from $6.25\times10^{-7}$ to $8\times10^{-5}$ across
configurations.

\paragraph{Evaluation.} We evaluate with 16 rollouts per problem at temperature 0.6 and
top-$p$ 0.95, a budget of 4{,}096 tokens and the LoRA adapter merged into the base model.

\paragraph{Measuring noise and statistical significance.} Each of the 84 combinations of
configuration, sampler and learning rate has two seeds, and the mean absolute difference
between the two seeds is 0.90 points. For normally distributed noise with a standard
deviation of $\sigma$ per run, the expected absolute difference between two runs is
$2\sigma/\sqrt{\pi}\approx1.13\,\sigma$, so the measured 0.90 corresponds to $\sigma=0.80$
points for a single run. A two-seed mean has a standard deviation of $\sigma/\sqrt{2}$, so the
difference between two two-seed means has a standard deviation of $\sigma=0.80$ and a 95\%
range of $\pm1.96\,\sigma=\pm1.56$. We call this range the noise range. At a single
configuration, the gains at $\shape{8}{8}$ ($+2.01$) and $\shape{32}{4}$ ($+2.07$) exceed the
noise range. A separate set of three seeds per method at $\shape{8}{8}$ with a rate of
$10^{-5}$ gives a gain of $+1.73$ after 300 steps, close to the $+1.59$ that the main runs
give at this rate. Smaller differences at a single configuration are not statistically
significant on their own, so claims across configurations use pooled tests. A pooled test
averages the gain over $n$ configurations and compares this mean with its standard error,
which is $0.80/\sqrt{n}$. For the mean gain over the ten configurations, the standard error is
$0.80/\sqrt{10}=0.25$.

\paragraph{The entropy relationship.} Over the ten configurations, the fitted line of
Section~\ref{sec:diversity} is gain $=5.12\,H-5.82$, where $H$ is the within-step entropy of
the bins in bits, averaged over the 250 steps of each run. When $\log_2P$ is added as a second
predictor, the coefficient of entropy is 5.83 and the coefficient of $\log_2P$ is $-0.15$,
which is not distinguishable from zero. Alone, $\log_2P$ gives $R^2=0.44$.

\paragraph{Compute.} Each run uses one H100 GPU. Training for 250 steps takes a median of
1.4 hours at 16 rollouts per step, 2.5 at 32, 4.5 at 64 and 8.6 at 128. The 168 runs of the
main comparison used about 850 GPU-hours of training. At $\shape{8}{8}$, a run of uniform sampling, TSC, static headroom, offline filtering or SPaCe
takes 4.4 to 4.8 hours. A DAPO run takes 7.8 hours, because it generates about twice as many
rollouts.

\small
\bibliographystyle{unsrtnat_initials}
\bibliography{refs}

\end{document}